\documentclass[romanappendices,journal,compsoc]{IEEEtran}

\usepackage[numbers]{natbib}
\usepackage{times}
\usepackage{subfigure}
\usepackage{graphicx}
\usepackage{amsmath,amssymb}
\usepackage{bbold}
\usepackage[breaklinks=true,colorlinks,bookmarks=false]{hyperref}
\usepackage{ragged2e}
\usepackage{multirow}
\usepackage[table,xcdraw]{xcolor}

\graphicspath{{Figures/}}

\begin{document}

\title{BEE-NET: A deep neural network to identify in-the-wild {B}odily {E}xpression of {E}motions}

\author{Mohammad Mahdi~Dehshibi~\IEEEmembership{Member, IEEE} and         
        David~Masip~\IEEEmembership{Senior~Member, IEEE}
\IEEEcompsocitemizethanks{\IEEEcompsocthanksitem \protect
Corresponding author: M. M. Dehshibiis affiliated with the Department of Computer Science and Engineering at Universidad Carlos III de Madrid in Legan\'{e}s, Spain (e-mail: mohammad.dehshibi@yahoo.com).}
\thanks{Manuscript \#TPAMI received XYZ; revised XYZ.}}

\IEEEtitleabstractindextext{
\justify
\begin{abstract}
In this study, we investigate how environmental factors, specifically the scenes and objects involved, can affect the expression of emotions through body language. To this end, we introduce a novel multi-stream deep convolutional neural network named BEE-NET. We also propose a new late fusion strategy that incorporates meta-information on places and objects as prior knowledge in the learning process. Our proposed probabilistic pooling model leverages this information to generate a joint probability distribution of both available and anticipated non-available contextual information in latent space. Importantly, our fusion strategy is differentiable, allowing for end-to-end training and capturing of hidden associations among data points without requiring further post-processing or regularisation. To evaluate our deep model, we use the Body Language Database (BoLD), which is currently the largest available database for the Automatic Identification of the in-the-wild Bodily Expression of Emotions (AIBEE). Our experimental results demonstrate that our proposed approach surpasses the current state-of-the-art in AIBEE by a margin of 2.07\%, achieving an Emotional Recognition Score of 66.33\%.
\end{abstract}

}

\maketitle
\IEEEraisesectionheading{\section{Introduction}\label{sec:introduction}}
\IEEEPARstart{H}{umans} recognise and perceive the emotional expressions of others to use these critical cues for successful nonverbal communication. Equipping computers with the ability to recognise, perceive, process, and simulate human affects will aid in the development of empathetic devices that can be used in monitoring certain mental/physical disorders~\cite{ashtari2022multi,dehshibi2023pain}, home-assistant devices, public safety, or the analysis of social media data~\cite{dehshibi2021deep,erol2019toward}. Emotion perception has typically been studied by analysing facial expressions~\cite{pons2020multitask}, body postures and gestures~\cite{luo2020arbee,noroozi2018survey}, and physiological signs~\cite{aghaahmadi2013clustering,dehshibi2010new,dehshibi2019cubic}. However, psychological studies have shown that body language can convey individuals’ emotional state~\cite{abramson2020social,aviezer2012body}, and the lack of a high-quality and diverse database with ground truth makes \textbf{A}utomatic \textbf{I}dentification of the in-the-wild \textbf{B}odily \textbf{E}xpression of \textbf{E}motions (AIBEE) a challenging research topic.

Several studies have assessed the variables that humans naturally use from a young age to provide a more reliable measure of an individual's emotional state by fusing different modalities such as facial expressions, speech prosody, and context-related data~\cite{kosti2020context,poria2017review}. Body language has recently been used to help understand other people's emotional states. According to Beck's cognitive depression theory~\cite{beck1967depression}, people suffering from depression tend to view themselves in mostly negative and dark environments. This intuition motivated us to propose a deep learning architecture that incorporates meta-information about the environment and the objects involved to reinforce AIBEE. 

We hypothesise that the environment and the objects involved greatly influence our perception of the in-the-wild bodily expressions of emotions. Therefore, rather than isolating human bodies in video frames, we propose a multi-stream convolutional neural network (BEE-NET) that incorporates prior knowledge about the joint probability of emotions and both available and anticipated non-available places/objects. This scalable architecture is differentiable, allowing for end-to-end model training without additional post-processing or regularisation. We evaluate the proposed method using the Body Language database (BoLD)~\cite{luo2020arbee}, which is by far the largest database available for AIBEE. Experimental results indicate that the proposed method outperforms the state-of-the-art in identifying discrete and continuous in-the-wild bodily expressions of emotions.

The rest of this paper is organised as follows: Section~\ref{sec:survey} surveys the previous studies. The proposed method is detailed in Section~\ref{sec:method}. Architectural design and implementation details are given in Section~\ref{sec:architectures}. Section~\ref{sec:experiment} presents experiments results. Finally, Section~\ref{sec:conclusion} concludes the paper.

\section{Literature review} \label{sec:survey}
Detecting actions in videos is a critical step toward understanding human behaviour. In this context, high-level reasoning is required not only in the spatial dimension but also in the temporal dimension. Simonyan and Zisserman~\cite{simonyan2014two} proposed one of the most promising multi-task convolutional neural network (CNN) architectures for action recognition, integrating spatial and temporal networks. Tran et al.~\cite{tran2018closer} proposed using 3D CNNs for general action recognition with the separate convolution of spatial and temporal filters, which increased recognition accuracy in AVA benchmark~\cite{gu2018ava}. Feichtenhofer et al.~\cite{feichtenhofer2019slowfast} suggested a two-pathway network with a low frame rate path focused on spatial information extraction and a high frame rate path focused on motion encoding. Hussein et al.~\cite{hussein2019timeception} developed a multi-scale temporal convolution approach, employing various kernel sizes and dilation rates to capture temporal dependencies.

Ulutan et al.~\cite{ulutan2020actor} proposed combining actor features with each Spatio-temporal region in the scene to generate attention maps between the actor and the context. Girdhar et al.~\cite{girdhar2019video} suggested a Transformer-style architecture for weighting actors based on contextual features. Wang and Gupta~\cite{wang2018non} suggested modelling a video clip to combine whole clip features and weighted proposal features computed by a graph convolutional network based on feature space similarities and Spatio-temporal distances between each detection. Tomei et al.~\cite{tomei2021video} propose a graph-based framework for learning high-level interactions between people and objects. The Spatio-temporal relationships are learned through self-attention on a multi-layer graph structure to link entities from consecutive clips for a wide range of Spatio-temporal dependencies.

Human action recognition in-the-wild has to deal with challenges such as different degrees of freedom, heterogeneity in people's behaviour, cluttered backgrounds, and variations in size, pose, and camera viewpoint~\cite{noroozi2018survey,yadav2021review}. Furthermore, existing benchmark databases, such as AVA~\cite{gu2018ava}, lacked tags for human affects. Therefore, several studies have focused on facial expressions rather than body gestures to identify emotions that are less subjective thanks to the introduction of the Facial Action Coding System~\cite{ekman1992argument} and more flexible as a result of the face having fewer degrees of freedom than the body~\cite{schindler2008recognizing}. Mollahosseini et al.~\cite{mollahosseini2016going} proposed a six-layer CNN with two convolution layers and four inception layers to classify facial expressions in still images. Pons and Masip~\cite{pons2020multitask} proposed a CNN committee in which a multi-task learning loss function integrates the detector of facial action units into the emotion learning model to solve the problem of learning multiple tasks with heterogeneously labelled data. Microexpressions were considered by Xu et al.~\cite{xu2017microexpression} to add nuances to facial expression recognition. Luvizon et al.~\cite{luvizon20182d} used a multi-task learning approach, and Li et al.~\cite{li2019spatio} used a Spatio-temporal graph CNN to leverage pose knowledge and improve the accuracy of bodily expression recognition across multiple modalities.

Given that bodily expressions can convey emotional states and, in some cases, are the only modality that can be used to correctly disambiguate the corresponding facial expression~\cite{abramson2020social,aviezer2012body,dehshibi2021deep}, recent studies attempted to incorporate bodily expressions of emotions into affective computing by introducing benchmark models and databases such as the EMOTIC~\cite{kosti2020context} and BoLD~\cite{luo2020arbee}. Kosti et al.~\cite{kosti2020context} introduced the EMOTIC database, which included information about valence, arousal, and dominance dimensions in addition to the six basic emotions. They used a two-stream CNN to extract features that represented the body expression and the scene description. Luo et al.~\cite{luo2020arbee} introduced the BoLD database and proposed a framework combining human identification, pose estimation, and learning representation to recognise bodily expressions of emotions. Kumar et al.~\cite{kumar2020noisy} propose using the BoLD database to train a noisy student network in which various face regions are processed independently using a multi-level attention mechanism to enhance facial expression recognition incrementally.

\section{BEE-NET Multi-stream Architecture} \label{sec:method}

We formulate the AIBEE as a regression problem in which the response $\mathbf{Y} \in \mathbb{R}^{d_{y}}$ is predicted given the input image $\mathbf{X} \in \mathbb{R}^{d_{x}}$, and the loss is measured by the mean squared error ($L_{2}$). Given a database of $n$ images $\mathcal{D} = \{(x_{i}, y_{i})\}_{i=1}^{n}$, with $x_{i} \in \mathbb{R}^{d_{x}}$ and $y_{i} \in \mathbb{R}^{d_{y}}$, our goal is to learn a neural network $\mathcal{H}(x) = \mathbb{E}[\mathbf{Y}|\mathbf{X} = x]$ that minimises the loss function.

\begin{figure}[!htb]
    \centering   
    \subfigure[]{\includegraphics[width=0.95\linewidth]{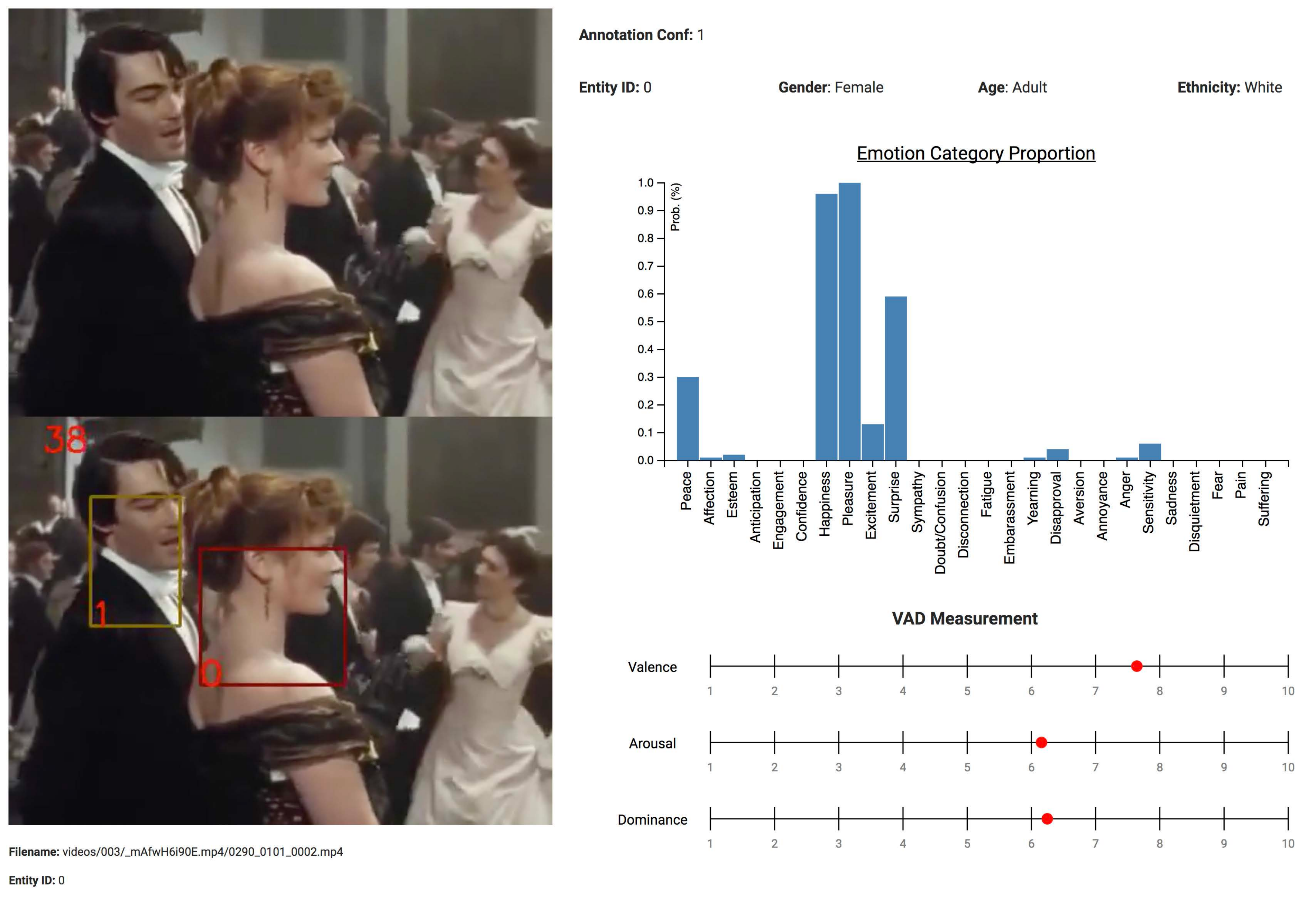}\label{fig:bold}}
    \subfigure[]{\includegraphics[width=0.577\linewidth]{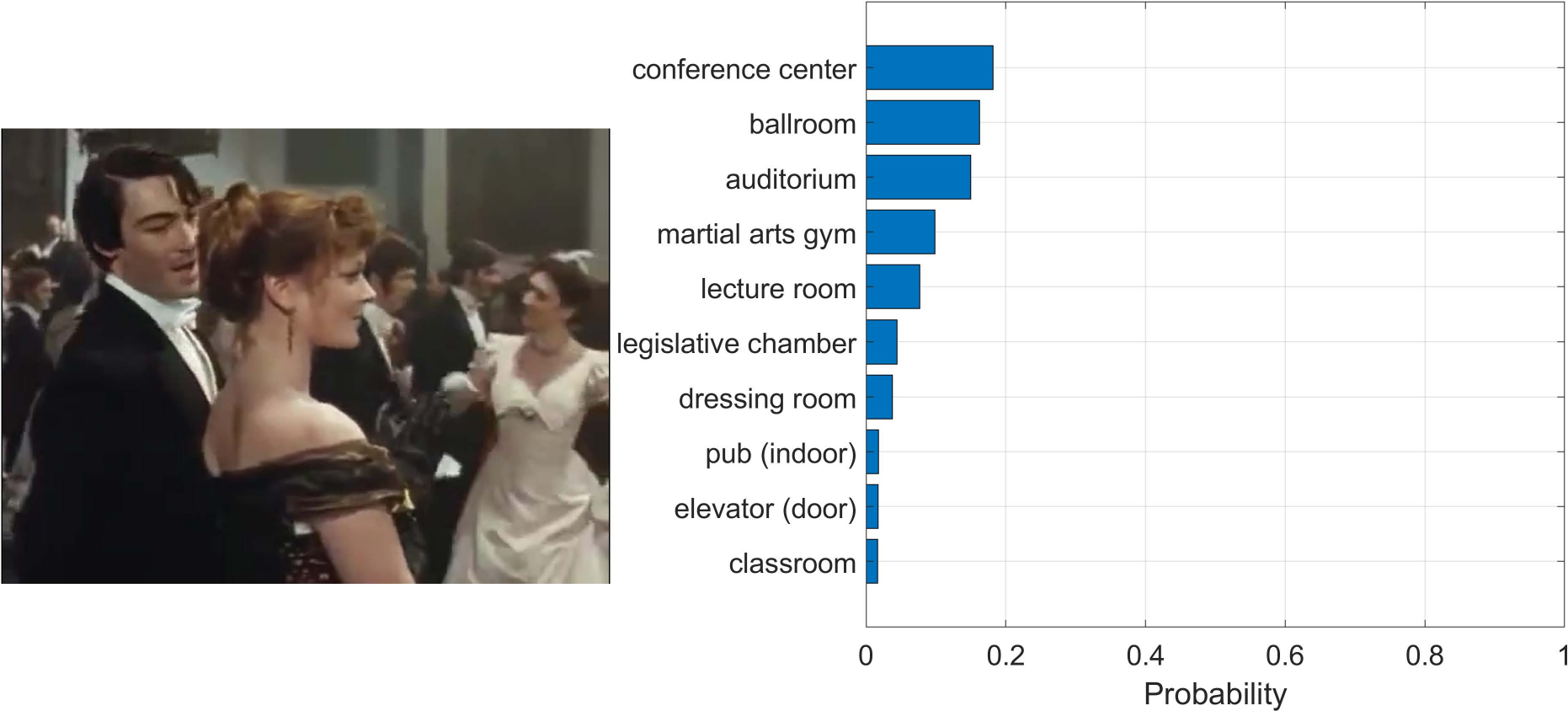}\label{fig:places}}
    \subfigure[]{\includegraphics[width=0.353\linewidth]{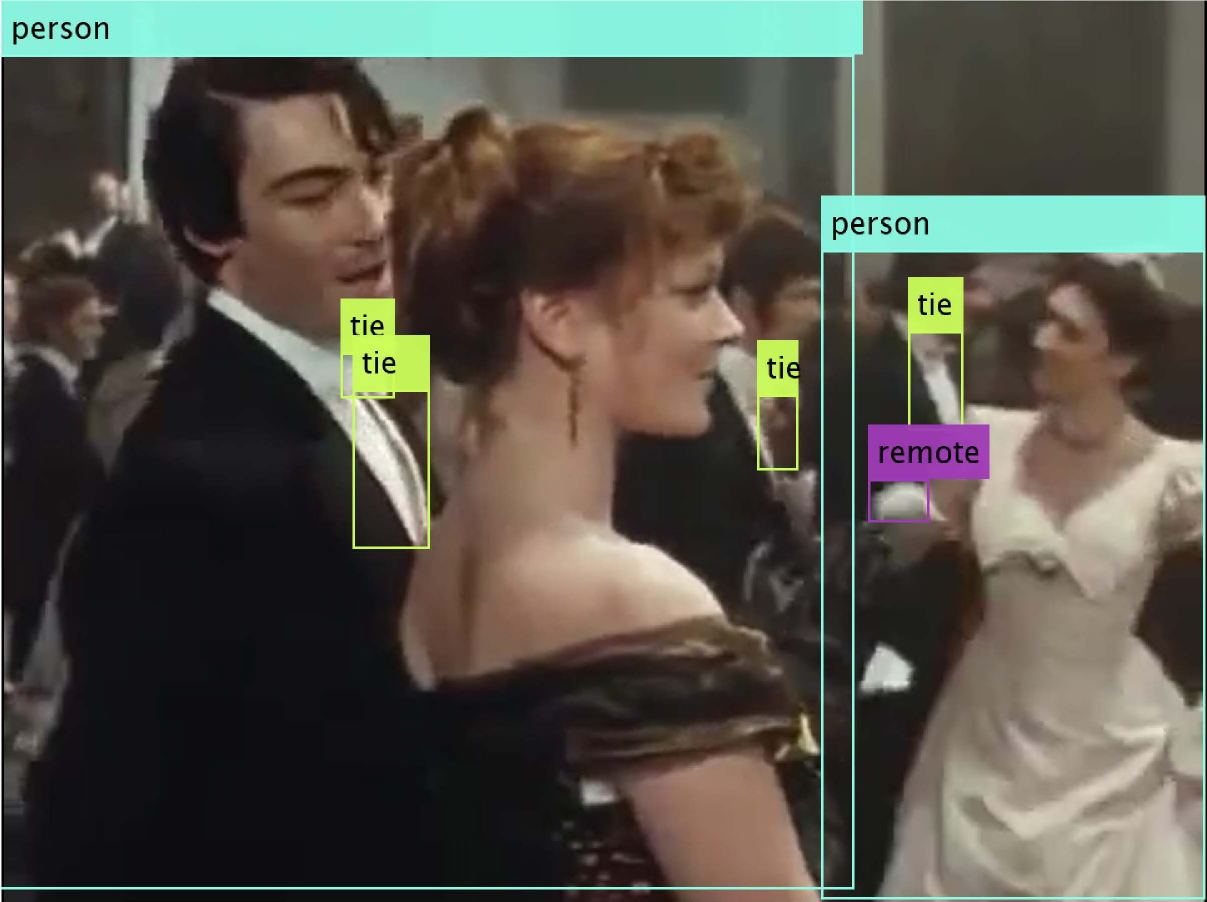}\label{fig:yolo}}
    \caption{(a) A sample from the BoLD database that mainly represents happiness. (b) Top-10 place tags were obtained by applying Places-CNN trained on the Places2 database to the input. (c) Object tags were obtained by applying YOLO trained on the Microsoft COCO database to the input.}
\label{fig:1}
\end{figure}

In this study, we use BoLD database~\cite{luo2020arbee} in which the input image is labelled by $d_{y} = 29$ emotions. We scale the input image $\mathbf{X}$ to the size of $d_{x} = 224 \times 224$. The emotion tag $\mathbf{Y}$ consists of 26 discrete emotions with values in the range of $[0,1]$ and three continuous emotions with values in the range of $[1,10]$ (see Fig.~\ref{fig:bold}). We assume that each image consists of three components, namely \textbf{emotion}, \textbf{place} and \textbf{object}. Emotion tags are provided in $\mathcal{D}$. To incorporate contextual information (\textit{place} and \textit{object}) that was not explicitly provided in the BoLD database, we used the transfer learning strategy to incorporate pseudo-ground truths into the learning model. Figures~\ref{fig:places} and~\ref{fig:yolo} show the provided place and object tags for the input image using the Places-CNN scene descriptor and the YOLO object detector~\cite{redmon2016you}, which trained on the Places2~\cite{zhou2018places} and Microsoft COCO~\cite{lin2014microsoft} databases, respectively.

Mensink et al.~\cite{mensink2014costa} reported that the co-occurrence of attributes in a model's training phase could occur with high probability in the testing phase. Integrating attribute co-occurrence also helps to diminish unwanted outliers introduced by domain integration and makes $L_{2}$ a good approximation for the loss function~\cite{gholami2020unsupervised,yu2019unsupervised}. However, fine-tuning the pre-trained models with a database where attribute labels are not explicitly provided is not possible.

\begin{figure*}[!htbp]
    \centering
    \subfigure[]{\includegraphics[width=0.74\linewidth]{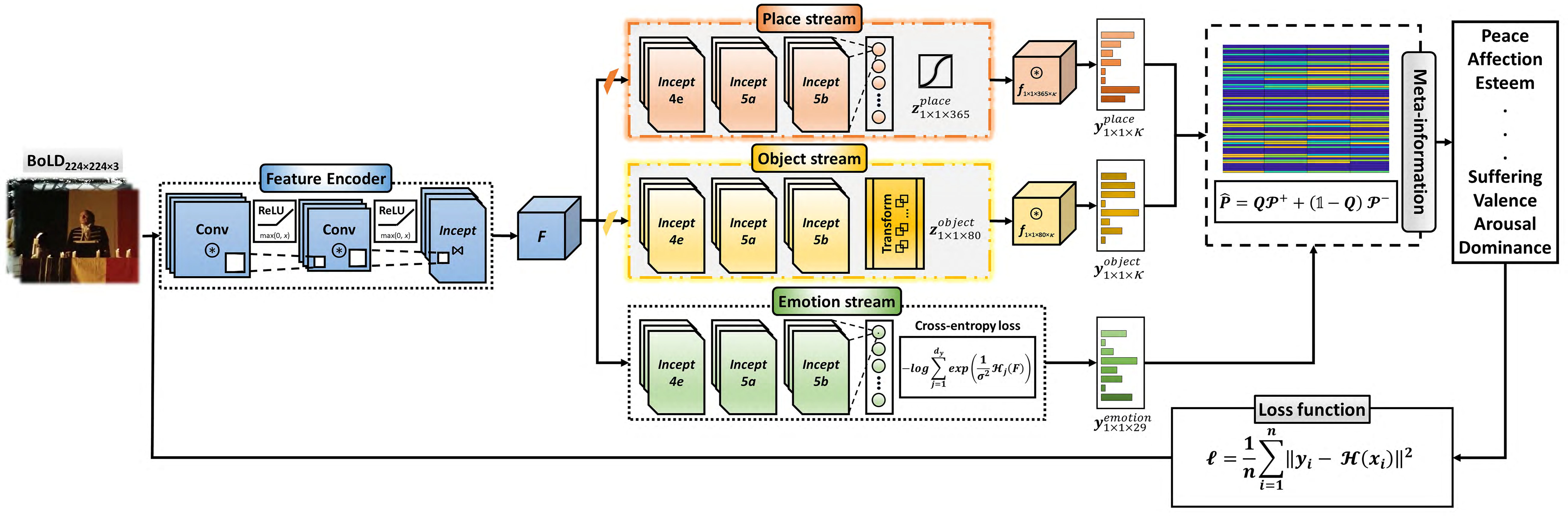}\label{fig:architecture}}
    \subfigure[]{\includegraphics[width=0.25\linewidth]{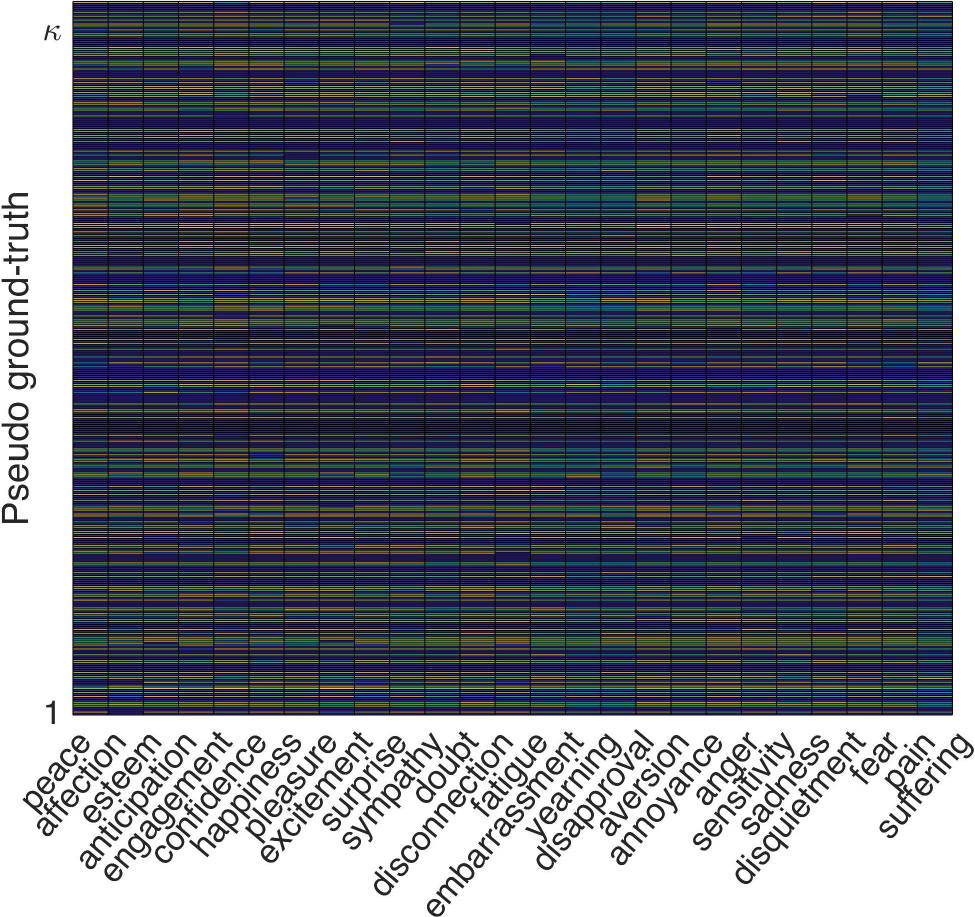}\label{fig:co-occure}}
    \caption{(a) BEE-NET architecture for the identification of in-the-wild bodily expression of emotions. Place and Object streams have a shade of grey in this schematic pipeline to highlight frozen layers at a zero learning rate during the training phase. (b) The pseudo-colour plot of the conditional probability of the \textit{emotion} tags given the \textit{place} and \textit{object} tags. The y-axis represents the probability of pseudo-ground-truth for the place and objects ($\kappa=365+80$), while the x-axis represents 26 discrete emotions. Note that the darker the blue, the lower the probability values.}
\end{figure*}

To address this problem, we propose a multi-stream convolutional neural network (BEE-NET) in which the pooling layer and loss function drive the emotion learning process during training using \textit{a priori} contextual information about the joint probability of emotions and both available and anticipated non-available places/objects. We formulated a derivable pooling scheme based on Bayesian theory to fuse the extracted uncertain information with the predicted image-based emotional states, allowing end-to-end model training to capture the hidden correlation between data without additional post-processing or regularisation. Fig.~\ref{fig:architecture} illustrates the proposed architecture.

\subsection{Network architecture} \label{sec:network}
The proposed architecture is composed of three main streams: (i) the scene descriptor stream, which determines the probability of place categories associated with the input image; (ii) the object detector stream, which detects objects involved in the input image; and (iii) the emotion stream, which focuses on learning the map between the input image and the emotions. The initial network stem $(\mathcal{H}^{base})$ is a feature encoder implemented with GoogLeNet~\cite{szegedy2015going} where the output of the Inception (4d) module $(F \in \mathbb{R}^{W \times H \times D})$ is fed into the rest of streams. $W=14$, $H=14$ and $D=528$ are the width, height, and number of channels of the feature map, respectively.

\noindent\textbf{(i) Scene descriptor \mbox{\boldmath $(\mathcal{H}^{place})$}}: resembles the architecture of the Places-CNN with the GoogLeNet backbone to provide the pseudo-ground-truth for place tags. The layers include the Inception (4e), (5a) and (5b), global average pooling, dropout, fully connected and softmax modules. More formally, $\mathcal{H}^{place}: F \rightarrow z^{place}$, where $z^{place} \in \mathbb{R}^{1 \times 1 \times 365}$ is a vector representing the probability of place categories. To incorporate $z^{place}$ into the proposed probabilistic model and harmonise the dimensionality of each stream, we convolve $z^{place}$ with a filter bank and add the bias using Eq.~\ref{eq:1}.

\begin{equation}
    \label{eq:1}
    y^{place} = z^{place} \circledast \mathbf{f}^{place} + b^{place}.
\end{equation}
where $\circledast$ performs the convolution, $\kappa$ is the number of output dimension, $\mathbf{f}^{place} \in \mathbb{R}^{1 \times 1 \times 365 \times \kappa}$ is the trainable filter, and $b^{place} \in \mathbb{R}^{\kappa}$ is the bias term.

\noindent\textbf{(ii) Object detector\mbox{\boldmath $(\mathcal{H}^{object})$}}: provides the pseudo-ground-truth for object tags. This stream is composed of three groups of serially connected convolution, ReLU, and batch normalisation layers, in which the entire topmost feature map is used to predict confidences for multiple categories of objects at a single stage. In Inception (4e), the filter size is set to $14 \times 14$ to match the number of channels in $F$. The second (5a) and third (5b) convolution layers have the filter size of $7 \times 7$ to enable the model to detect small objects. These layers are followed by the transform and output layers, respectively. The transform layer transforms raw CNN output into the form required for object detections and is followed by the output layer, which defines and implements the 7 anchor boxes and loss function used to train the detector. Anchor boxes extract the activations of the last convolutional layer and match predicted bounding boxes with the ground truth. More formally, $\mathcal{H}^{object}: F \rightarrow z^{object}$, where $z^{object} \in \mathbb{R}^{1 \times 1 \times 80}$ represents the probability of object categories. With the same purpose as in the scene descriptor stream, we convolve $z^{object}$ with a trainable filter bank $\mathbf{f}^{object} \in \mathbb{R}^{1 \times 1 \times 80 \times \kappa}$ and add it with bias $b^{object} \in \mathbb{R}^{\kappa}$ as in Eq.~\ref{eq:2}.
\begin{equation}
    \label{eq:2}
    y^{object} = z^{object} \circledast \mathbf{f}^{object} + b^{object}.
\end{equation}

\noindent\textbf{(iii) Emotion stream \mbox{\boldmath $(\mathcal{H}^{emotion})$}}: This stream learns a regression model that maps the output of the initial network stem $F$ into emotion ($y^{emotion} \in \mathbb{R}^{1 \times 1 \times d_{y}}$). This stream is mainly composed of the Inception (4e), (5a) and (5b), global average pooling, dropout, and fully connected modules. Inspired by~\cite{kendall2018multi}, we formulate the regression task with a softmax likelihood in Eq.~\ref{eq:3}. Note that here we intentionally drop the \emph{emotion} superscript to simplify mathematical notations.

\begin{equation}
    \label{eq:3}
    p\left ( y|\mathcal{H} (F), \sigma \right ) = \mathrm{softmax}\left ( \frac{1}{\sigma^{2}} \mathcal{H}(F) \right ).
\end{equation}
where $\sigma$ is a learnable noise scalar that observes the uniformity of a discrete distribution. In the maximum likelihood inference, we maximise the log-likelihood of the model. To obtain the log-likelihood of the model (Eq.~\ref{eq:3}), we can write the cross-entropy loss as Eq.~\ref{eq:4}, where $\mathcal{H}_{j}(F)$ is the $j^{\text{th}}$ element of the vector $\mathcal{H}(F)$.

\begin{align}
    \label{eq:4}
    \log p\left ( y=i|\mathcal{H} (F), \sigma \right )_{1 \leq i \leq d_{y}} = \nonumber \\ \frac{1}{\sigma^{2}} \mathcal{H}_{i}(F) - \log \sum_{j=1}^{d_{y}} \exp \left ( \frac{1}{\sigma^{2}} \mathcal{H}_{j}(F) \right ).
\end{align}

In the formulation of the cross-entropy loss, we assumed that $\sum_{j=1}^{d_{y}} \exp \left ( \frac{1}{\sigma^{2}} \mathcal{H}_{j}(F) \right ) \approx \left ( \sum_{j=1}^{d_{y}} \exp (\mathcal{H}_{j}(F)) \right )^{\frac{1}{\sigma^{2}}}$ to simplify the optimisation objective. This approximation becomes equality when $\sigma \longmapsto 1$. This loss function is differentiable and avoids any division by zero. It also prevents the weights of the \emph{emotion} stream from converging to zero when the co-occurrence probability of available and non-available meta-information is incorporated into the architecture.

\subsection{Probabilistic Pooling for Late Fusion in BEE-NET}
Due to the association between emotions and the diversity of objects/places, it is not easy to accurately estimate the conditional probabilities of a given image $x$ with regard to all attribute labels at the same time. To address this issue, we propose a multi-stream architecture (BEE-NET), which includes place ($y^{place}$) and object ($y^{object}$) auxiliary streams in addition to the emotion stream ($y^{emotion}$). As shown in Fig.~\ref{fig:architecture}, the lower layers ($F$) are shared across streams, while the top layers are separated to focus on the attributes for different contextual information. To fuse the extracted uncertain information from the place and object streams with the predicted image-based emotional states, we build a matrix, as given in Eq.~\ref{eq:05}, to incorporate prior knowledge about the joint probability of emotions and places/objects, which can be considered of as softmax classifier outputs stacked into a $2 \times d_{y}$ matrix $\mathcal{P}$. Note that we substituted $y^{place}$, $y^{object}$, and $y^{emotion}$ with with $\mathbb{A}_{1}$, $\mathbb{A}_{2}$ and $\mathbb{B}$, to simplify mathematical notations.
\begin{equation}
    \label{eq:05}
    \mathcal{P} = \begin{bmatrix}
    \mathrm{Pr}(\mathbb{B}_{1}|\mathbb{A}_{1}) & \mathrm{Pr}(\mathbb{B}_{2}|\mathbb{A}_{1}) & \cdots & \mathrm{Pr}(\mathbb{B}_{d_{y}}|\mathbb{A}_{1})\\
    \mathrm{Pr}(\mathbb{B}_{1}|\mathbb{A}_{2}) & \mathrm{Pr}(\mathbb{B}_{2}|\mathbb{A}_{2}) & \cdots & \mathrm{Pr}(\mathbb{B}_{d_{y}}|\mathbb{A}_{2})
    \end{bmatrix}.
\end{equation}
where $\mathrm{Pr}(\mathbb{B}_{i}|\mathbb{A}_{j})$ is the conditional probability of $\mathbb{B}_{i}$ given the feature of the $j^{\mathrm{th}},~ j \in \{1,2\}$ stream. To calculate $\mathrm{Pr}(\mathbb{B}_{i}|\mathbb{A}_{j})$ in the training set, we must first determine the number of occurrences of each emotion label as well as the number of co-occurrences of this emotion label given place/object pseudo-labels. If $N_{i}$ is the number of occurrences of the $i^{\mathrm{th}}$ emotion in the dataset, then $p_{i} = \mathrm{Pr}(\mathbb{B}_{i}) = \frac{N_{i}}{n}$ is the probability of the $i^{\mathrm{th}}$ label. Likewise, if $N_{i,j}$ is the co-occurrence number of the dataset's $i^{\mathrm{th}}$ and $j^{\mathrm{th}}$ labels, $\mathcal{C}_{i,j} = \mathrm{Pr}(\mathbb{B}_{i} \cap \mathbb{A}_{j}) = \frac{N_{i,j}}{n}$ is a matrix, representing the joint probability of $\mathbb{B}_{i}$ and $\mathbb{A}_{j}$. Therefore, we can obtain the conditional probability of one attribute given another, as shown in Eq.~\ref{eq:06}. Fig.~\ref{fig:co-occure} shows the pseudo-colour plot of $\mathcal{P}^{+}$ discovered on the BoLD training set, where the darker the blue, the lower the probability values.

\begin{equation}
    \label{eq:06}
    \mathcal{P}^{+}_{i,j} = \mathrm{Pr}(\mathbb{B}_{i}|\mathbb{A}_j)=\frac{\mathrm{Pr}(\mathbb{B}_{i} \cap \mathbb{A}_{j})}{\mathrm{Pr}(\mathbb{A}_{j})}=\frac{\mathcal{C}_{i,j}}{p_{j}}.
\end{equation}
where $p_{j}$ is a high-order posterior probability expressing a correlation between the place ($\mathbb{A}_{1}$) and object ($\mathbb{A}_{2}$) streams given the $i^{\mathrm{th}}$ emotion, as calculated by Eq.~\ref{eq:07}.

\begin{align}
    \label{eq:07}
    \mathrm{Pr}\left ( \mathbb{A}_{j=1}^{i} \right ) = \mathrm{Pr}\left ( \mathbb{A}_{j=1}^{i}|\mathbb{A}_{j=2}^{i,1},\mathbb{A}_{j=2}^{i,2},\cdots,\mathbb{A}_{j=2}^{i,\kappa} \right ),\nonumber\\
    \mathrm{Pr}\left ( \mathbb{A}_{j=2}^{i} \right ) = \mathrm{Pr}\left ( \mathbb{A}_{j=2}^{i}|\mathbb{A}_{j=1}^{i,1},\mathbb{A}_{j=1}^{i,2},\cdots,\mathbb{A}_{j=1}^{i,\kappa} \right ).
\end{align}

Because these high-order posterior probabilities cannot be calculated precisely, we approximate $p_{j}$ using local max-pooling over streams and apply local max-pooling over $\mathcal{P}^{+}_{i,j}$ (see Eq.~\ref{eq:08}).

\begin{equation}
    \label{eq:08}
    Q_{1 \times d_{y}} \approx \max_{j \in \{1,2\}} \mathrm{Pr}(\mathbb{B}_{i}|\mathbb{A}_{j}) = \max_{j \in \{1,2\}} \mathcal{P}^{+}_{i,j}.
\end{equation}

The prediction result of the auxiliary streams, which are extracted from the input image using empirical knowledge about places/objects, has a significant impact on the accuracy of Eq.~\ref{eq:08}. Because these attributes are correlated, it is difficult for the streams to adequately separate them, resulting in a biased estimation of the conditional probability vector using Eq.~\ref{eq:08}. In addition, the columnar elements of $\mathrm{Pr}(\mathbb{B}_{i}\cap\mathbb{A}_{j})$ are extremely small, resulting in relatively small max-pooling values. On the other hand, the unavailability of a meta-information in an input image may lead to the availability of another meta-information or vice versa. For instance, it is impossible to imagine people playing soccer in a pool full of water. For this reason, we also use conditional probability to measure the probability of the anticipated non-available meta-information ($\mathcal{P}^{-}$) by Eq.~\ref{eq:09}.

\begin{equation}
    \label{eq:09}
    \begin{split}
        \mathcal{P}^{-}_{i,j} = \mathrm{Pr}(\mathbb{B}_{i}|\neg \mathbb{A}_{j}) = & \frac{\mathrm{Pr}(\mathbb{B}_{i} \cap \neg \mathbb{A}_{j})}{\mathrm{Pr}(\neg \mathbb{A}_{j})} \\ 
        = & \frac{\mathrm{Pr}(\mathbb{B}_{i})-(\mathbb{B}_{i} \cap \mathbb{A}_{j})}{1-\mathrm{Pr}(\mathbb{A}_{j})} = \frac{p_{i}-\mathcal{C}_{i,j}}{1-p_{j}}
    \end{split}
\end{equation}

We then use $\mathcal{P}^{+}$ and $\mathcal{P}^{-}$ matrices to calculate the joint probability $\hat{\mathcal{P}}$ as in Eq.~\ref{eq:10}, which is the output of the proposed pooling scheme.

\begin{equation}
\label{eq:10}
\hat{\mathcal{P}} = Q\mathcal{P}^{+} + (\mathbb{1}-Q)\mathcal{P}^{-}.
\end{equation}
where $\mathbb{1} \in \mathbb{R}^{1 \times d_{y}}$ is a full one vector. By adding the proposed pooling scheme to the rest of the architecture, the assembled network $\mathcal{H}$ can predict bodily expressions of emotion $\tilde{y}$ given the input image $x$ by Eq.~\ref{eq:11}.

\begin{equation}
    \label{eq:11}
    \tilde{y} = \mathcal{H}(x) = \frac{1}{\lambda}\hat{\mathcal{P}}.
\end{equation}
where $\lambda$ is used to regulate the prediction results of the streams. In order to minimise the loss function ($\ell$) in Eq.~\ref{eq:12}, we use the stochastic gradient descent algorithm combined with the backpropagation strategy to update the network parameters.

\begin{equation}
    \label{eq:12}
    \ell = \frac{1}{n} \sum_{i=1}^{n}||y_{i} - \tilde{y}_{i}||^{2}.
\end{equation}


\section{Architecture details} \label{sec:architectures}
Experiments were performed on the NVIDIA RTX 2080 GPU with 8 GB of memory in MATLAB 2020a using the Deep Learning toolbox. The parameters for the proposed architecture are as follows:

\noindent\textbf{-- Scene descriptor stream}: In $(\mathcal{H}^{place})$, we set weights and parameters of layers to the identical ones in Places-CNN, which was trained with Places2~\cite{zhou2018places} database. To prevent layers of this stream from being overfitted during the training of $\mathcal{H}^{emotion}$ on the BoLD database and force the stream to determine the probability of place categories, we set the learning rates to zero.

\noindent\textbf{-- Object detector stream}: In $\mathcal{H}^{object}$, we replace pooling layers with $16 \times 16$ and $32 \times 32$ strides, respectively. We use 7 anchor boxes $(width,~height) \in \{$(10,~13), (16,~30), (33,~23), (30,~61), (62,~45), (59,~119), (116,~90)$\}$ with an intersection over union threshold of 0.4. We set the non-maximal suppression (NMS) threshold to 0.4 in order to keep the best bounding box. Since we use Microsoft COCO database~\cite{lin2014microsoft}, each anchor box $(x,y,w,h,s,c)$ has 85 properties, where $(x,y,w,h)$ represents bounding box properties, $s$ is the detection score and $c$ is an array whose size is equal to the number of classes. Anchor boxes are only added to the final stream's layer in which each cell contains $7 \times 85 = 595$ elements making $8 \times (7 \times 7) \times 392$ predictions.

\begin{figure*}[!htbp]
    \centering
    \subfigure[]{\includegraphics[width=0.32\linewidth]{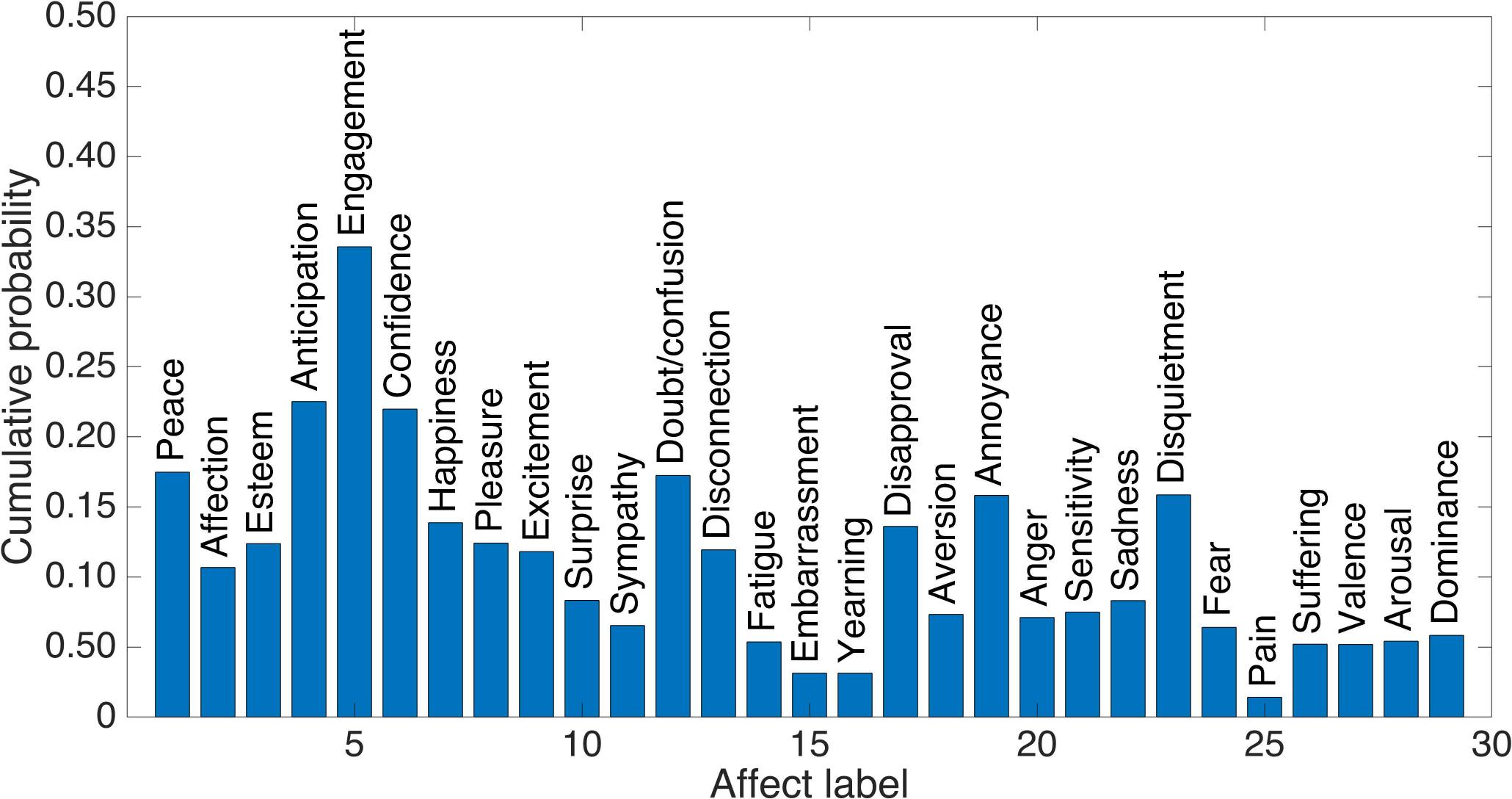} \label{fig:partition}}
    \subfigure[]{\includegraphics[width=0.32\linewidth]{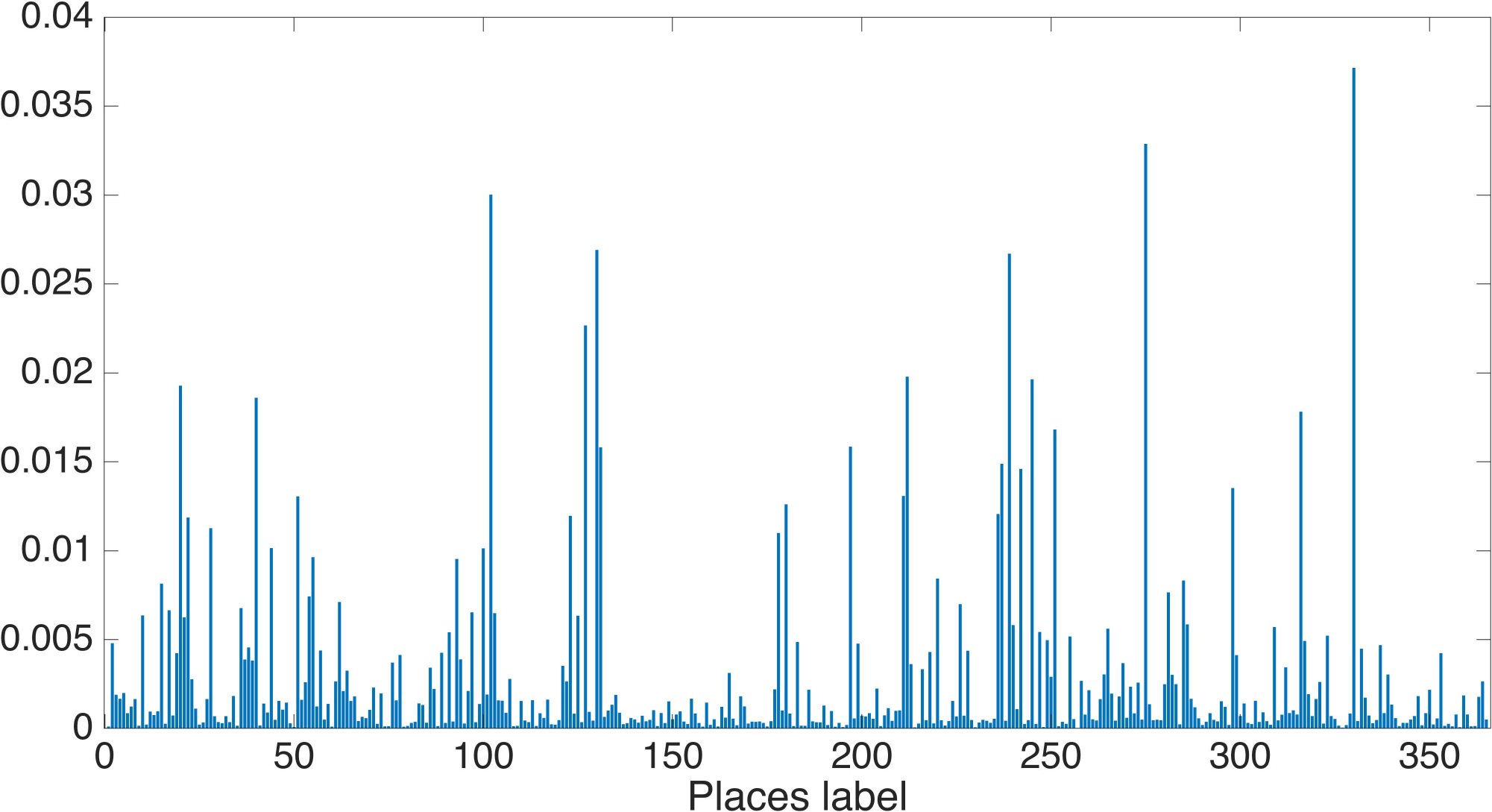} \label{fig:Pdistribution}}
    \subfigure[]{\includegraphics[width=0.32\linewidth]{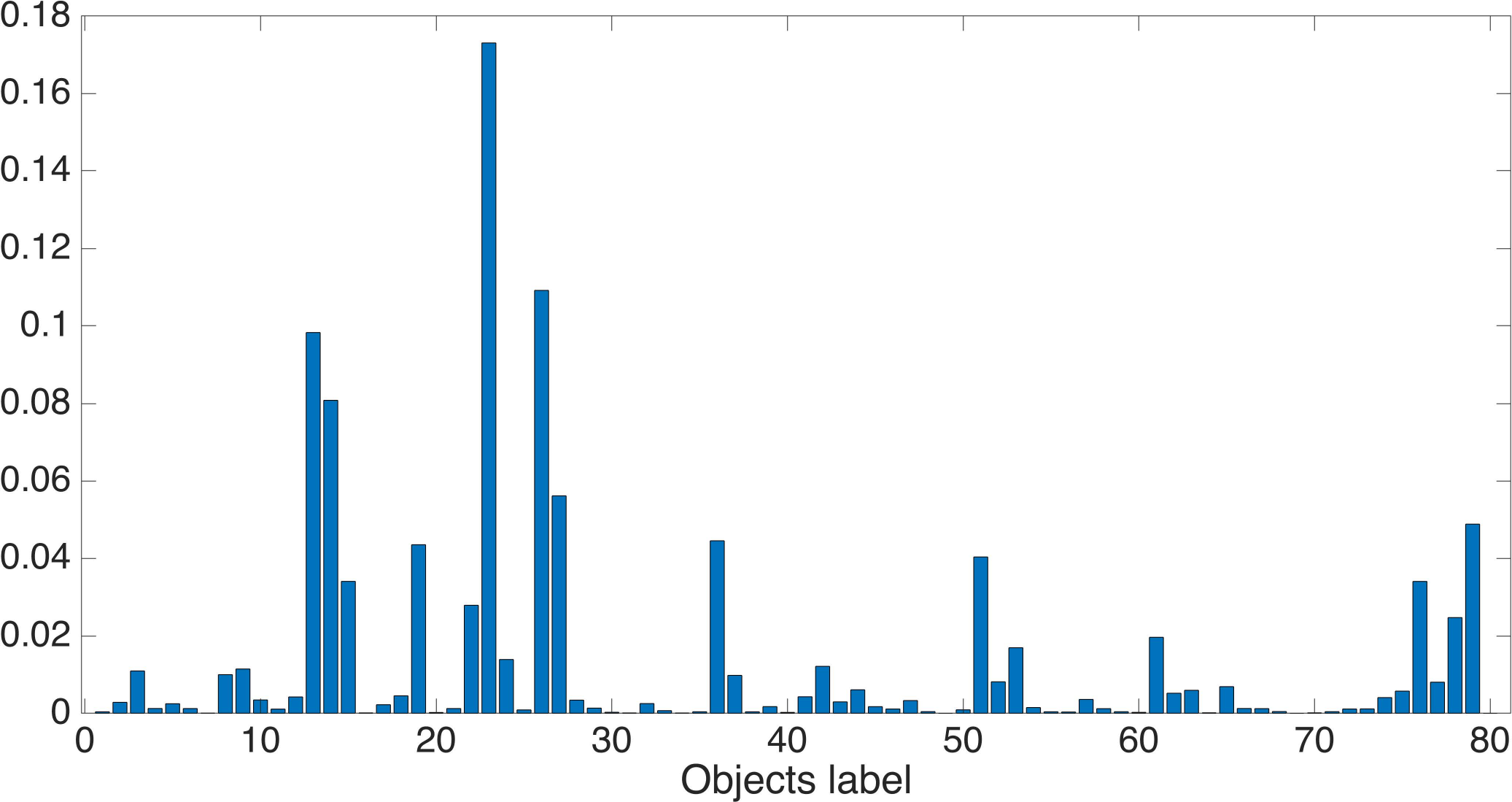}\label{fig:Odistribution}}
    \caption{Cumulative probability of (a) labels in BoLD database, (b) pseudo-tag provided by applying place-CNN to BoLD database, and (c) pseudo-tag provided by applying YOLO object detector to BoLD database.}
\end{figure*}

To set the weights and parameters of $\mathcal{H}^{object}$, we first built a deep neural network that was the serial connection of $\mathcal{H}^{base}$ and $\mathcal{H}^{object}$. We trained this network in a separate scenario with the Microsoft COCO database~\cite{lin2014microsoft}. Throughout the training, we use a batch size of 8, a momentum of 0.9, and a decay of $5 \times 10^{-3}$. The learning rate is set to $10^{-2}$, $10^{-3}$ and $10^{-4}$ for the first 75 epochs, the next 30 epochs, and the final 30 epochs, respectively. This policy is used to prevent the model from diverging due to unstable gradients. After training, we transferred the weights and parameters of the layers that had the same architecture as $\mathcal{H}^{object}$ to the proposed architecture. Finally, we set the learning rates in these layers to zero in order to prevent these layers from being overfitted during the training of $\mathcal{H}^{emotion}$ on the BoLD database.

\noindent\textbf{-- Emotion stream}: In $\mathcal{H}^{emotion}$, we train the network using stochastic gradient descent with momentum 0.9. The initial learning rate is set to $10^{-2}$, and the learning rate is decreased by a factor of 0.1 every 45 epochs. We set the maximum number of training epochs to 90 and use a mini-batch with 8 observations at each iteration, where the training data is shuffled before each training epoch. The training parameters for the initial network stem ($\mathcal{H}^{base}$) are set to the same values. 

The proposed multi-stream architecture has two hyper-parameters $\kappa$ and $\lambda$ that are used to align the meta-information dimensionality in the calculation of $y^{place}$ and $y^{object}$ and to regulate the prediction results of the streams in Eq.~\ref{eq:11}. To find the appropriate value for $\kappa$, we apply the pre-trained YOLO object detector and Places-CNN to the training set. By applying Places-CNN, the input image ($x_{i}$) maps into a 365-dimensional vector ($z_{i}^{place}$) containing the probability of place tags. However, this map cannot be straightforwardly built for the YOLO object detector. 

The list of potential objects for each input image ($x_{i}$) may contain a different number of elements. Also, this list may contain multiple instances from one object with different confidence scores. To unify the codomain to which the input image is mapped, we define a vector with 80 entries ($z_{i}^{object}$), each of which corresponds to one object tag. We then assign the normalised cumulative confidence score of the detected objects to the corresponding entries of the object list ($z^{object}$) and set the rest of the entries to zero. For example, 2 instances of `person', 5 instances of `tie' and 1 instance of `remote' were detected by the object detector in Fig.~\ref{fig:yolo}. The normalised cumulative confidence scores of 0.19, 0.12 and 0.01 are assigned to the corresponding entries for these objects in the output list, and the rest of the entries are set to 0.

The maximum value of $\kappa$ is equal to the minimum dimensionality of $\mathcal{H}^{place}$ and $\mathcal{H}^{object}$ outputs, \emph{i.e.}, $\kappa = 80$. We calculate the normalised sum of outputs for the \emph{place} ($\overline{z}^{place} = \frac{1}{n}\sum_{i=1}^{n}z_{i}^{place}$) and \emph{object} ($\overline{z}^{object} = \frac{1}{n}\sum_{i=1}^{n}z_{i}^{object}$) tags in the training set to find the minimum value of $\kappa$. The distribution of $\overline{z}^{place}$ and $\overline{z}^{object}$ are shown in Figures~\ref{fig:Pdistribution} and~\ref{fig:Odistribution}, respectively. By applying the threshold of 0.01 to $\overline{z}^{place}$ and $\overline{z}^{object}$, we found that 27 \emph{place} and 14 \emph{object} tags can meet the threshold. Therefore, we set the minimum value of $\kappa$ to 14 and test its impact by changing the value from 14 to 80 with the step of 6. 

\begin{figure}[!htbp]
    \centering
    \includegraphics[width=0.9\linewidth]{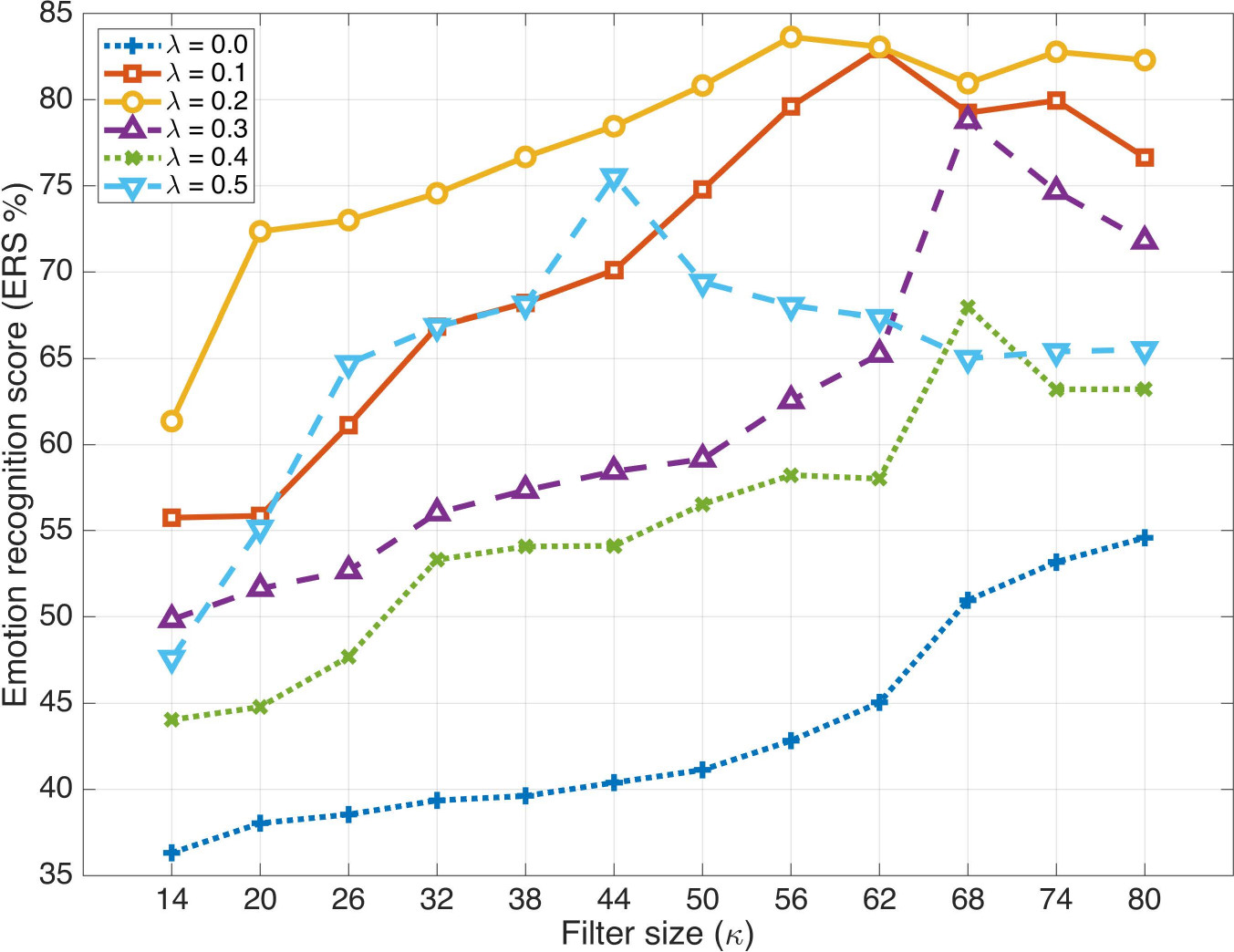}
    \caption{The emotion recognition score for the proposed architecture as a function of $(\kappa, \lambda)$. The best trade-off between these two hyper-parameters is $(\kappa, \lambda) = (56, 0.2)$, resulting in an ERS value of 83.64\%.}
    \label{fig:ers_kappa_lambda}
\end{figure}

The parameter $\kappa$ enables $\mathbf{f}^{place}$ and $\mathbf{f}^{object}$ filters in Eq.~\ref{eq:1} and~\ref{eq:2} to select a subset of relevant predictions of $\mathcal{H}^{place}$ and $\mathcal{H}^{object}$ streams that contribute the most to the calculation of Eq.~\ref{eq:05}--\ref{eq:10}. The trainable filter kernel also enables the architecture to learn the correspondences of the feature maps for meta-information to minimise the loss function in Eq~\ref{eq:12}. Another hyper-parameter is $\lambda$ that we evaluate its impact by changing the value from 0 to 0.5 with a 0.1 step. We trained the proposed architecture as a function of $(\kappa, \lambda)$ and plotted the emotion recognition score (ERS) to find the best trade-off between these two hyper-parameters (see Fig.~\ref{fig:ers_kappa_lambda}). We obtained an ERS value of 83.64 at $\kappa = 56$ and $\lambda = 0.2$ and used these values in the rest of the experiments.

From Fig.~\ref{fig:ers_kappa_lambda} and the analysis of prediction errors, we inferred that larger values of $(\kappa, \lambda)$ not only increase computational complexity but also make the result ($\tilde{y}$) more inclined to the meta-information estimation of $\mathcal{P}^{+}$ and $\mathcal{P}^{-}$.

\section{Experiments} \label{sec:experiment}
\subsection{Body Language database}
We performed our experiments on the Body Language database (BoLD)~\cite{luo2020arbee}, which contains by far the most data for AIBEE. The videos in BoLD were chosen from the publicly available AVA database~\cite{gu2018ava}, which includes a collection of YouTube movie IDs. There are 9,876 video clips of humans expressing emotion, mainly through body gestures. The crowd-sourcing platform was employed to annotate the database with two widely accepted emotional categorisations. 

There are 26 labels for categorical emotions, including \{Peace, Affection, Esteem, Anticipation, Engagement, Confidence, Happiness, Pleasure, Excitement, Surprise, Sympathy, Doubt/confusion, Disconnection, Fatigue, Embarrassment, Yearning, Disapproval, Aversion, Annoyance, Anger, Sensitivity, Sadness, Disquietment, Fear, Pain, Suffering\}.  The continuous emotional dimensions are Valence, Arousal, and Dominance. It should be noted that, while the gathered videos are annotated based on body language, the movies with a close-up of the face rather than the entire or partially-occluded body remain unlabelled. 

\subsection{Evaluation metrics and experimental protocols} 
We use the mean $R^{2}$ score (Eq.~\ref{eq:13}) to evaluate the proposed regression model. The $R^{2}$ metric calculates the ratio of explained variance ($y$) to measure how well the unseen samples are likely to be predicted by the model's independent variables.

\begin{equation}
    \label{eq:13}
    R^{2}(y,\tilde{y}) = 1- \frac{\sum_{i=1}^{n}(y_{i}-\tilde{y}_{i})^{2}}{\sum_{i=1}^{n}(y_{i}-\varepsilon)^{2}}, ~~ \varepsilon = \frac{1}{n}\sum_{i=1}^{n}y_{i}.
\end{equation}
where $\tilde{y}_{i}$ is the is the predicted value of the $i$-th sample, and $y_{i}$ is the corresponding true value. 

Since we formulated the AIBEE as a regression problem, we applied the $\max$ threshold to convert the predicted quantity of the regression model into discrete buckets for emotions. The use of $\arg\max$ lets us evaluate the efficiency of the proposed regression model using Precision, Recall and $F1$, where $F1$ is a weighted average of the precision and recall metrics. We report the average precision (m$AP$), the mean area under the receiver's operating characteristic curve (m$RA$), and the mean of $F1$ for the assessment. For ease of comparison, the Emotion Recognition Score (ERS) is also defined in Eq.~\ref{eq:14}.

\begin{align}
    \label{eq:14}
    \mathcal{G} = \{R^{2}, mAP, mRA\}, ~ \Delta = R^{2} + \left (\frac{mAP+mRA}{2} \right), \nonumber \\
    \text{ERS} = \frac{\Delta - \min(\mathcal{G}_{i})}{\max(\mathcal{G}_{i}) - \min(\mathcal{G}_{i})},~ 1 \leq i \leq 3.
\end{align}

In the experiments, we did not use data augmentation techniques to prevent unrealistic changes in colour, angle and position that could alter our hypothesis about the relationship between the representation of emotions and the environment and the objects involved. We partitioned the database into train, test and validation sets containing 60\%, 30\% and 10\% of samples, respectively. The members of each set were chosen at random in such a way that the distribution of BoLD for each set was observed (see Fig.~\ref{fig:partition}).


\subsection{Experimental results} \label{sec:results}
Table~\ref{tbl:1} shows the performance of the proposed architecture along with competitive methods~\cite{carreira2018action,luo2020arbee,wang2016temporal} to verify the effectiveness of the BEE-NET. Following~\cite{luo2020arbee}, we used a random method based on priors (referred to as `Chance') as a basis for comparison. Laban Movement Analysis (LMA)~\cite{von1966choreutics} was originally developed to characterize dance movements by a set of structural and physical characteristics through representing body, effort, form and space. Because of the proximity of body language to this representation, Luo et al.~\cite{luo2020arbee} used LMA to identify the bodily expression of emotions. They used the method proposed by Cao et al.~\cite{cao2019openpose} to detect 2D poses in still images to use in LMA and reported promising results on the BoLD database for AIBEE.

We also compared the proposed architecture with two CNN architectures that were considered as state-of-the-art in action recognition. To use the Temporal Segment Networks (TSN)~\cite{wang2016temporal} in AIBEE, we split each video into 2 segments. During the training stage, one frame is randomly sampled for each segment, and the classification result is averaged over all the sampled frames. We set the learning rate and batch size to $10^{-3}$ and 16, respectively. Other training requirements are similar to the original version. The two-stream inflated 3D CNN~\cite{carreira2018action} uses 3D convolution to learn Spatio-temporal features in an end-to-end way. However, we replaced 3D convolution with 2D convolution to perform experiments on still images that were randomly sampled from each video. In our experiment, we set the learning rate and batch size to $10^{-2}$ and 16, respectively. We preserved other training details, as stated in the original version.

\begin{table}[!htbp]
\centering
\caption{The performance of BEE-NET in the test set. The mean measurement of continuous emotions is $R^{2}$. The percentage metrics of m$AP$, m$RA$ and m$F1$ are used to report the performance of classifying discrete emotions. ERS is used to report the emotion identification score. Note that `Chance' refers to a random method based on priors.}
\label{tbl:1}
\begin{tabular}{llllll}
\hline
\multirow{2}{*}{Method} & \multirow{2}{*}{ERS (\%)} & Regression & \multicolumn{3}{c}{Classification (\%)} \\ \cline{3-6} 
 &  & m$R^{2}$ & m$AP$ & m$RA$ & m$F1$ \\ \hline
Chance & 61.75 & 0 & 11.75 & 50 & 19.02 \\
Luo et al.~\cite{luo2020arbee} & 64.08 & 0.0947 & 17.48 & 62.59 & 27.32 \\
Wang et al.~\cite{wang2016temporal} & 62.24 & 0.0760 & 14.02 & 57.65 & 22.55 \\
Carreira et al.~\cite{carreira2018action} & 64.26 & 0.1007 & 17.33 & 61.2 & 27.01 \\
\textbf{BEE-NET} & \textbf{66.33} & \textbf{0.1493} & \textbf{23.18} & \textbf{71.56} & \textbf{35.01} \\ \hline
\end{tabular}
\end{table}

\begin{figure*}[!htbp]
    \centering
    \includegraphics[width=0.95\linewidth]{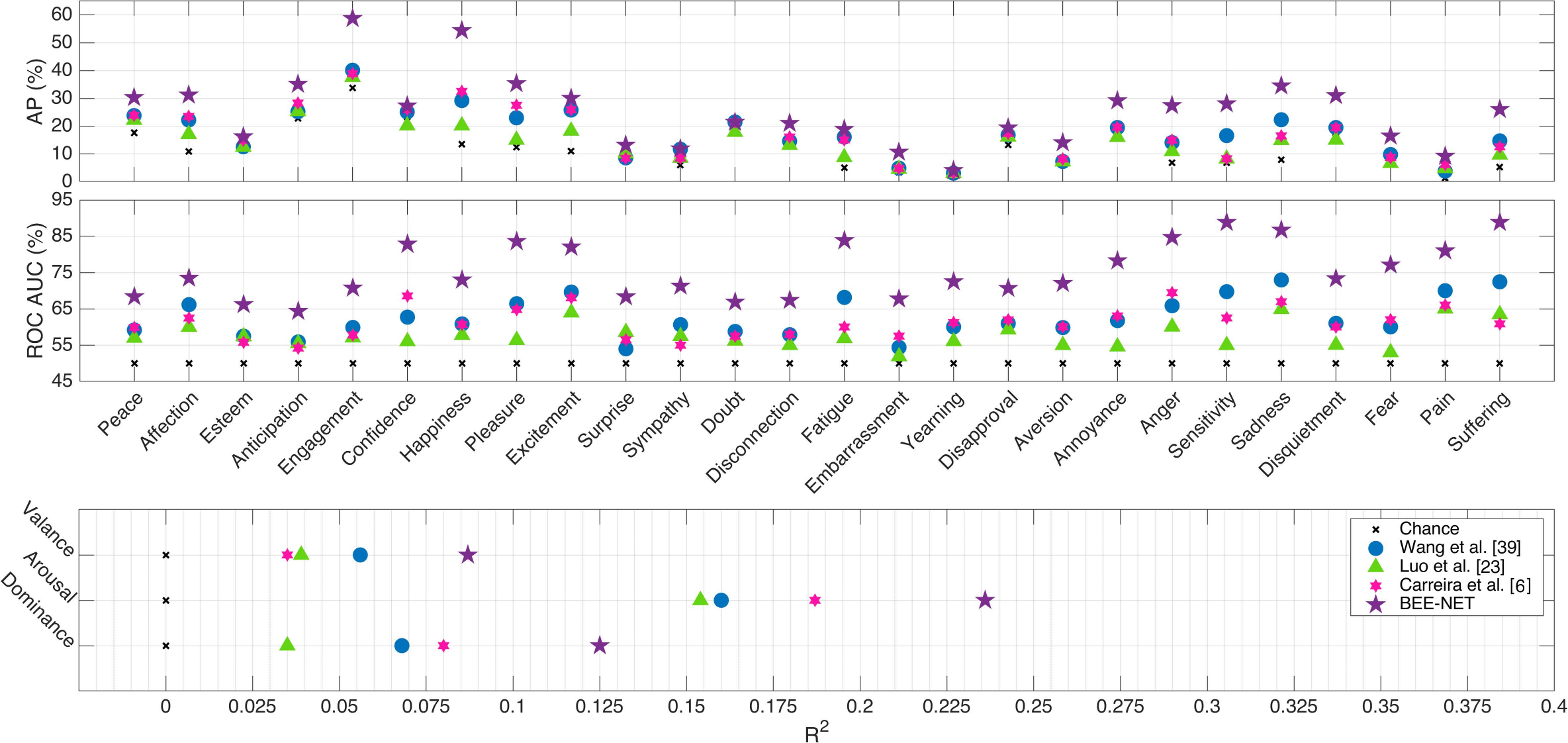}
    \caption{Classification performance for discrete emotions is reported based on the average precision (AP) in the [first row] and area under the receiver's operating characteristic curve (RA) in the [second row]. The regression performance for continuous emotions is reported on the basis of the $R^{2}$ score in the [third row].}
    \label{fig:metrics}
\end{figure*}

Figure~\ref{fig:metrics} provides comprehensive metric comparisons of all methods of each categorical and dimensional emotion. From Table~\ref{tbl:1} and Fig.~\ref{fig:metrics}, it could be said that our hypothesis regarding the influence of the environment and the object involved in the disclosure of human emotions is valid. Moreover, it can be seen in Fig.~\ref{fig:metrics} that the \{Engagement, Happiness, Pleasure, Anticipation, Sadness\} categories comprise the top-5 predictions. Indeed, the proposed architecture could appropriately address the bias problem (see Fig.~\ref{fig:partition}) towards \{Engagement, Anticipation, Confidence, Peace, Doubt\} in the BoLD database. In the Ablation study, we will demonstrate that the formulation of the pooling scheme, where the probability of both available and anticipated non-available items being considered, could contribute to this achievement. 

In the assessment of continuous emotions, all methods show a greater performance of arousal regression than valence and dominance. However, compared to the subjective test reported in~\cite{luo2020arbee}, humans showed better valence-recognition performance than arousal. This distinction between human and model output indicates that domain knowledge and experience in other contexts allow humans to make better decisions in completely new situations.

\subsection{Ablation study}
In the ablation study, we conducted four sets of experiments to better understand the efficacy of the BEE-NET for the AIBEE task. We therefore examine:

\noindent\textbf{--- Contribution of pre-trained model weights}: In this experiment, instead of assigning random weights to the $\mathcal{H}^{base} + \mathcal{H}^{emotion}$ network filters, we used GoogLeNet filters' weight that was trained with ImageNet~\cite{szegedy2015going}, Places2 and Microsoft COCO. We reduced the training steps to 45 iterations and retained all the other parameters as described in Section~\ref{sec:architectures}. The findings in Table~\ref{tbl:2} show that the initialisation of filters with random weights leads to a marginally better performance of AIBEE.

\begin{table}[!htbp]
\centering
\caption{Ablation study on the effect of pre-trained models.}
\label{tbl:2}
\begin{tabular}{llllll}
\hline
 &  & Regression & \multicolumn{3}{c}{Classification (\%)} \\ \cline{3-6} 
\multirow{-2}{*}{Initial weight} & \multirow{-2}{*}{ERS (\%)} & m$R^{2}$ & m$AP$ & m$RA$ & m$F1$ \\ \hline
\multicolumn{6}{l}{\cellcolor[HTML]{EFEFEF}Ablation study} \\ \hline
\textbf{Random} & \textbf{65.42} & \textbf{0.1241} & \textbf{21.37} & \textbf{69.82} & \textbf{32.72} \\
ImageNet & 65.09 & 0.1007 & 19.89 & 66.36 & 30.60 \\
Places2 & 64.59 & 0.1103 & 18.72 & 64.66 & 29.03 \\
Microsoft COCO & 64.44 & 0.0997 & 18.36 & 64.02 & 28.53 \\ \hline
\multicolumn{6}{l}{\cellcolor[HTML]{EFEFEF}Proposed architecture} \\ \hline
BEE-NET & 66.33 & 0.1493 & 23.18 & 71.56 & 35.01 \\ \hline
\end{tabular}
\end{table}

\noindent\textbf{--- Contribution of \mbox{\boldmath $\mathcal{H}^{place}$} and \mbox{\boldmath $\mathcal{H}^{object}$}}: In two experiments, we eliminated $\mathcal{H}^{place}$ and $\mathcal{H}^{object}$ streams and trained the altered architecture ($\mathcal{H} \circleddash \mathcal{H}^{place}$ and $\mathcal{H} \circleddash \mathcal{H}^{object}$) with BoLD data to understand the effect of each stream. In this experiment, we retained all other parameters as described in Section~\ref{sec:architectures}, except for $\kappa$. The value $\kappa$ was respectively set to 100 and 40 during the operation of the $\mathcal{H}^{place}$ and $\mathcal{H}^{object}$ streams. The results of this ablation study are shown in Table~\ref{tbl:3}.

\begin{table}[!htbp]
\centering
\caption{Ablation study on the effect of $\mathcal{H}^{place}$ and $\mathcal{H}^{object}$ streams.}
\label{tbl:3}
\begin{tabular}{llllll}
\hline
 &  & Regression & \multicolumn{3}{c}{Classification (\%)} \\ \cline{3-6} 
\multirow{-2}{*}{Architecture} & \multirow{-2}{*}{ERS (\%)} & m$R^{2}$ & m$AP$ & m$RA$ & m$F1$ \\ \hline
\multicolumn{6}{l}{\cellcolor[HTML]{EFEFEF}Ablation study} \\ \hline
$\mathcal{H} \circleddash \mathcal{H}^{object}$ & 64.91 & 0.0804 & 18.85 & 63.56 & 29.07 \\
$\mathcal{H} \circleddash \mathcal{H}^{place}$ & 63.55 & 0.0589 & 15.23 & 56.48 & 23.99 \\ \hline
\multicolumn{6}{l}{\cellcolor[HTML]{EFEFEF}Proposed architecture} \\ \hline
BEE-NET & 66.33 & 0.1493 & 23.18 & 71.56 & 35.01 \\ \hline
\end{tabular}
\end{table}

It can be inferred from Table~\ref{tbl:3} that the impact of $\mathcal{H}^{place}$ is greater than that of $\mathcal{H}^{object}$. This effect is also evident in Table~\ref{tbl:2}, where the use of pre-trained model filters with the Microsoft COCO database, among other databases, resulted in lower performance. Furthermore, the frames were primarily sourced from older movies within the BoLD database, which inherently possess lower quality and smaller sizes. Therefore, locating the right and appropriate objects in the scene is met with an error that later propagates to the architecture. For example, a ``remote" object in Fig.~\ref{fig:yolo} was detected by the object detector, which is an incorrect and unrelated object to the context. Moreover, considering the intent of the BoLD database, `Person' is the dominant object in all scenes. Therefore, the majority of $z^{object}$ entries have a value of zero that, despite applying the $\mathbf{f}^{object}$ kernel, the sparsity propagates to $y^{object}$ feature vector. In this way, the classifier must deal with the sparse representation in the latent space that reduces its performance. However, the influence of $\mathcal{H}^{object}$ in the proposed architecture is undeniable as its combination with $\mathcal{H}^{place}$ could improve the state-of-the-art in identifying in-the-wild bodily expressions of emotions by 2.07\%.

\noindent\textbf{--- Contribution of face to AIBEE}: To examine the impact of the face on BEE-NET performance, we filled the face area with black pixels in all database frames. Then, we trained the proposed architecture with new images, where the network parameters are retained as described in Section~\ref{sec:architectures}. Although masking the face had little effect on `Person' detection due to the presence of $\mathcal{H}^{object}$ stream, the ERS metric decreased to 62.35\%. This remarkable decrease emphasises how facial expression can support the bodily expression of emotions.

\noindent\textbf{--- Contribution of fusion strategies}: Incorporating fusion strategies into deep learning models that deal with multi-modal input or extract features using multi-stream architectures can improve accuracy and performance. These fusion strategies are typically categorised into early, intermediate, and late fusion categories. This ablation study aims to compare the efficacy of the proposed probabilistic pooling-based late fusion strategy with other fusion strategies to demonstrate how leveraging meta-information can outperform conventional fusion strategies. To do this, we have modified the proposed architecture in the following ways and present the results in Table~\ref{tbl:4}.

\begin{enumerate}
    \item The early fusion strategy involves merging and processing all input data at the beginning of the neural network before performing feature extraction. Although an early fusion strategy can be useful for simple tasks, it may lead to overfitting. In this ablation study, we cannot apply the early fusion strategy as we feed the BEE-NET with uni-modal data.
    \item The intermediate fusion strategy in multi-stream deep models combines features from multiple streams at an intermediate layer before performing the classification or regression. This fusion strategy usually uses concatenation, element-wise addition, or element-wise multiplication. In this ablation study, we replaced the proposed probabilistic pooling-based fusion strategy with an intermediate fusion strategy in BEE-NET. Specifically, we concatenated the output features of the place ($\mathcal{H}^{place}$) and object ($\mathcal{H}^{object}$) streams with the features of the initial network stem ($\mathcal{H}^{base}$). We then passed the concatenated features through a fully connected layer to obtain a fused feature vector. Finally, we fed the emotion stream ($\mathcal{H}^{emotion}$) with the fused features and continued the forward pass.
    \item The late fusion strategy in multi-stream deep neural networks involves combining the outputs of multiple streams at a later stage in the network architecture, typically after the individual streams have been processed by their own set of layers. In this ablation study, we conducted two experiments to assess the proposed fusion strategy. In the first experiment, we removed the probability of anticipated non-available meta-information by eliminating the $\mathcal{P}^{-}$ and $\hat{\mathcal{P}}$ terms from Eq.~\ref{eq:10} and trained the BEE-NET with $\hat{\mathcal{P}} = Q \mathcal{P}^{+}$. In the second experiment, we substituted the proposed fusion strategy with the one proposed by Kendall et al.~\cite{kendall2018multi} and proceeded with the forward pass.
\end{enumerate}

\begin{table*}[!htbp]
    \caption{Ablation study results on fusion strategies' contribution, measured by mutual information (MI) and entropy (E) metrics. MI increases with accuracy and dependence, while E increases with uncertainty and randomness in model predictions.}
    \label{tbl:4}
    \centering
    \resizebox{0.70\linewidth}{!}{%
        \begin{tabular}{lcllllll}
        \hline
         &  & \multicolumn{1}{c}{Regression} & \multicolumn{3}{c}{Classification} & \multicolumn{2}{c}{Uncertainly} \\ \cline{3-8} 
        \multirow{-2}{*}{Fusion method} & \multirow{-2}{*}{ERS (\%)} & m$R^{2}$ & m$AP$ & m$RA$ & m$F1$ & E~\textdownarrow & MI~\textuparrow \\ \hline
        \multicolumn{8}{l}{\cellcolor[HTML]{EFEFEF}Ablation study} \\ \hline
        Early fusion & N/A & N/A & N/A & N/A & N/A & N/A & N/A \\
        Intermediate fusion & \textbf{64.50} & 0.1095 & \textbf{17.46} & 60.68 & 25.86 & 3.32 & 2.95 \\
        Late fusion ($\hat{\mathcal{P}} = Q \mathcal{P}^{+}$) & 63.42 & \textbf{0.1104} & 16.70 & \textbf{62.75} & 26.37 & \textbf{1.98} & \textbf{3.81} \\
        Late fusion (Kendall et al.~\cite{kendall2018multi}) & 63.77 & 0.1036 & 17.14 & 62.70 & \textbf{26.92} & 2.03 & 3.66 \\ \hline
        \multicolumn{8}{l}{\cellcolor[HTML]{EFEFEF}Proposed architecture} \\ \hline
        BEE-NET & 66.33 & 0.1493 & 23.18 & 71.56 & 35.01 & 1.25 & 4.51 \\ \hline
        \end{tabular}%
    }
\end{table*}

Table~\ref{tbl:4} reveals that the intermediate fusion strategy exhibits a marginal improvement over both late fusion strategies regarding the evaluation metrics. However, intriguingly, the late fusion strategy from which the probability of anticipated non-available meta-information is excluded shows reduced uncertainty in terms of Entropy (see Eq.~\ref{eq:15}) and Mutual Information (see Eq.~\ref{eq:16})\footnote{Entropy quantifies the uncertainty of a single random variable, while Mutual Information measures the shared information between two random variables. In our experiment,  we employ kernel density estimation with a Gaussian kernel (i.e., $N(\mu, \sigma) = N(0,1)$) to estimate the probability density functions for predicted and ground-truth values. Subsequently, we derive the joint and marginal probability density functions from the estimated ones. In our experiment, we set the bandwidth value $h$ to $1.06 \sigma n^{-\frac{1}{5}}$~\cite{silverman1998density} to minimise the mean integrated squared error, where $n=29$ represents the number of predicted values.}.

\begin{equation}
    \label{eq:15}
    \mathrm{E} = -\sum_{\tilde{y}\in \tilde{\mathbf{Y}}}p(\tilde{y})\log \left ( p(\tilde{y}) \right ),
\end{equation}
where $\tilde{\mathbf{Y}}$ is the predicted values, and $p(\tilde{y})$ is the probability of $\tilde{\mathbf{Y}}$ taking the value $\tilde{y}$.
\begin{equation}
    \label{eq:16}
    \mathrm{MI} = \sum_{\tilde{y} \in \tilde{\mathbf{Y}}}\sum_{y \in \mathbf{Y}}P_{(\mathbf{Y},\tilde{\mathbf{Y}})}(y,\tilde{y})\log\left( \frac{P_{(\mathbf{Y},\tilde{\mathbf{Y}})}(y,\tilde{y})}{P_{\mathbf{Y}}(y)P_{\tilde{\mathbf{Y}}}(\tilde{y})} \right ).
\end{equation}
where $P_{(\mathbf{Y},\tilde{\mathbf{Y}})}$ is the joint probability density function of $\mathbf{Y}$ and $\tilde{\mathbf{Y}}$, and $P_{\mathbf{Y}}$ and $P_{\tilde{\mathbf{Y}}}$ are the marginal probability density functions of 
$\mathbf{Y}$ (i.e., ground-truth values) and $\tilde{\mathbf{Y}}$, respectively.

These findings support our hypothesis that the proposed probabilistic pooling-based late fusion strategy effectively utilises the meta-information provided by the place and object streams to confidently identify bodily expressions of emotions. Our results also highlight that the intermediate fusion strategy tends to dilute individual modalities' strengths and combine the individual models' uncertainties. This can lead to an overall higher level of uncertainty and inferior performance compared to our proposed fusion strategy. This observation aligns with the results presented in Table~\ref{tbl:3}, where we showed that if the object stream detects incorrect, unrelated, and dominant objects concerning the nature of the database, it can increase the sparsity, noise, and outlier in the latent space. These defects can be propagated throughout the model when the fusion strategy fails to leverage the correlation between the outputs from different streams.

Finally, it is essential to note that the decision between late and intermediate fusion strategies should be carefully evaluated based on the specific requirements of the task, the characteristics of the data, and the available computational resources. While late fusion strategies may offer advantages in specific scenarios, intermediate fusion strategies can provide opportunities for more efficient and effective integration of \emph{multi-modal} information. It is imperative to consider the unique contributions of each modality and the potential impact of noise when designing multi-modal deep learning architectures for optimal performance.
\section{Conclusion} \label{sec:conclusion}
Humans rely on emotional expressions to create meaningful interpersonal relationships. To enable computers to recognise, perceive, interpret, and simulate emotions as humans do, they must be equipped with the ability to understand and simulate human affects. Recent research has attempted to integrate bodily expressions of emotions into affective computing, as bodily expressions can convey emotional states and are sometimes the only modality that can accurately disambiguate the corresponding facial expression.

The present study investigated how environmental and object factors may influence the perception of in-the-wild bodily expressions of emotions. We proposed a novel multi-stream convolutional neural network (BEE-NET), which integrates pre-trained place and object recognition networks to represent contextual information. To incorporate this information, we formulated a derivable pooling scheme based on Bayes' theorem, which fuses the extracted uncertain information with the predicted image-based emotional states. This allows for end-to-end model training and the acquisition of \textit{a priori} information on the joint probability of emotions and both available and anticipated non-available places/objects, driving the emotion learning process during training.

Our experimental results, obtained using the Body Language Database (BoLD), the largest database available for identifying in-the-wild bodily expressions of emotions, demonstrate that our proposed method outperforms the state-of-the-art in identifying categorical (discrete) and continuous in-the-wild bodily expressions of emotions. Specifically, we validated our hypothesis that explicitly incorporating the co-occurrences of available and anticipated non-available places/objects into the fusion strategy can simplify and guide the learning process, removing the need for the network to automatically discover the impact of these relationships on the decision.

Overall, our proposed method, BEE-NET, provides an efficient and effective approach to incorporating contextual information into the emotion recognition process, which can lead to improved performance in real-world applications.


\section*{Acknowledgements}

\noindent \textbf{Funding:} This research was supported by a grant from the Spanish Ministry of Science, Innovation, and Universities (RTI2018-095232-B-C22) and the NVIDIA Hardware grant program. This work was also supported by the European Research Council (ERC) through the Horizon 2020 research and innovation program under grant agreement number 101002711. Mohammad Mahdi~Dehshibi received partial funding from this source.\\

\noindent \textbf{Author Contributions:} Mohammad Mahdi¬Dehshibi contributed to the conception of the idea, defined the scope of the study, and conducted the experiments. All authors participated in the discussion of results, provided feedback on the manuscript, and assisted in writing and editing.\\

\noindent \textbf{Competing Interests:} The authors declare no competing interests.

{\small
\bibliographystyle{IEEEtran.bst}
\bibliography{references}

\begin{thebibliography}{10}\itemsep=-1pt

\bibitem{abramson2020social}
L. Abramson, R. Petranker, I. Marom, and H. Aviezer.
\newblock {Social interaction context shapes emotion recognition through body
  language, not facial expressions}.
\newblock {\em Emotion}, 21(3), 2020.

\bibitem{aghaahmadi2013clustering}
M. Aghaahmadi, M.~M. Dehshibi, A. Bastanfard, and M. Fazlali.
\newblock Clustering persian viseme using phoneme subspace for developing
  visual speech application.
\newblock {\em Multim. Tools Appl.}, 65(3):521--541, 2013.

\bibitem{ashtari2022multi}
Mona Ashtari-Majlan, Abbas Seifi, and Mohammad~Mahdi Dehshibi.
\newblock {A Multi-Stream Convolutional Neural Network for Classification of
  Progressive MCI in Alzheimer’s Disease Using Structural MRI Images}.
\newblock {\em {IEEE Journal of Biomedical and Health Informatics}},
  26(8):3918--3926, 2022.

\bibitem{aviezer2012body}
H. Aviezer, Y. Trope, and A. Todorov.
\newblock {Body Cues, Not Facial Expressions, Discriminate Between Intense
  Positive and Negative Emotions}.
\newblock {\em Science}, 338(6111):1225--1229, 2012.

\bibitem{beck1967depression}
A.~T. Beck.
\newblock {\em Depression: Clinical, experimental, and theoretical aspects}.
\newblock Hoeber Medical Division, Harper \& Row, 1967.

\bibitem{cao2019openpose}
Z. Cao, G. Hidalgo, T. Simon, S.~E. Wei, and Y. Sheikh.
\newblock {OpenPose: Realtime Multi-Person 2D Pose Estimation Using Part
  Affinity Fields}.
\newblock {\em IEEE Trans. Pattern Anal. Mach. Intell.}, 43(1):172--186, 2021.

\bibitem{carreira2018action}
J. Carreira and A. Zisserman.
\newblock {Quo Vadis, Action Recognition? A New Model and the Kinetics
  Dataset}.
\newblock In {\em CVPR}, pages 4724--4733. IEEE, 2017.

\bibitem{dehshibi2021deep}
M.~M. Dehshibi, Bita Baiani, Gerard Pons, and David Masip.
\newblock {A Deep Multimodal Learning Approach to Perceive Basic Needs of
  Humans From Instagram Profile}.
\newblock {\em {IEEE Trans. Affect.}}, 14(2):944--956, 2023.

\bibitem{dehshibi2010new}
M.~M. Dehshibi and A. Bastanfard.
\newblock A new algorithm for age recognition from facial images.
\newblock {\em {Signal Process}}, 90(8):2431--2444, 2010.

\bibitem{dehshibi2023pain}
M.~M. Dehshibi, T.~A. Olugbade, F. Diaz-de Maria, N. Bianchi-Berthouze, and A.
  Tajadura-Jim{\'e}nez.
\newblock {Pain Level and Pain-Related Behaviour Classification Using GRU-Based
  Sparsely-Connected RNNs}.
\newblock {\em {IEEE Journal of Selected Topics in Signal Processing}},
  17(3):677--688, 2023.

\bibitem{dehshibi2019cubic}
M.~M. Dehshibi and J. Shanbehzadeh.
\newblock {Cubic norm and kernel-based bi-directional PCA: toward age-aware
  facial kinship verification}.
\newblock {\em {Vis. Comput.}}, 35(1):23--40, 2019.

\bibitem{ekman1992argument}
P. Ekman.
\newblock An argument for basic emotions.
\newblock {\em Cogn. Emot.}, 6(3-4):169--200, 1992.

\bibitem{erol2019toward}
B.~A. Erol, A. Majumdar, P. Benavidez, P. Rad, K.~K.~R. Choo, and M. Jamshidi.
\newblock {Toward Artificial Emotional Intelligence for Cooperative Social
  Human--Machine Interaction}.
\newblock {\em IEEE Trans. Comput. Soc. Syst.}, 7(1):234--246, 2020.

\bibitem{feichtenhofer2019slowfast}
C. Feichtenhofer, H. Fan, J. Malik, and K. He.
\newblock {SlowFast Networks for Video Recognition}.
\newblock In {\em ICCV}, pages 6201--6210. IEEE, 2019.

\bibitem{gholami2020unsupervised}
B. Gholami, P. Sahu, O. Rudovic, K. Bousmalis, and V. Pavlovic.
\newblock {Unsupervised Multi-Target Domain Adaptation: An Information
  Theoretic Approach}.
\newblock {\em IEEE Trans. Image Process.}, 29:3993--4002, 2020.

\bibitem{girdhar2019video}
R. Girdhar, J. Carreira, C. Doersch, and A. Zisserman.
\newblock {Video Action Transformer Network}.
\newblock In {\em CVPR}, pages 244--253. IEEE, 2019.

\bibitem{gu2018ava}
C. Gu, C. Sun, D.~A. Ross, C. Vondrick, C. Pantofaru, Y. Li, S.
  Vijayanarasimhan, G. Toderici, S. Ricco, R. Sukthankar, C. Schmid, and J.
  Malik.
\newblock {AVA: A Video Dataset of Spatio-Temporally Localized Atomic Visual
  Actions}.
\newblock In {\em CVPR}, pages 6047--6056. IEEE, 2018.

\bibitem{hussein2019timeception}
N. Hussein, E. Gavves, and A.~W.~M. Smeulders.
\newblock {Timeception for Complex Action Recognition}.
\newblock In {\em CVPR}, pages 254--263. IEEE, 2019.

\bibitem{kendall2018multi}
A. Kendall, Y. Gal, and R. Cipolla.
\newblock {Multi-task Learning Using Uncertainty to Weigh Losses for Scene
  Geometry and Semantics}.
\newblock In {\em CVPR}, pages 7482--7491. IEEE, 2018.

\bibitem{kosti2020context}
R. Kosti, J. Alvarez, A. Recasens, and A. Lapedriza.
\newblock {Context Based Emotion Recognition Using EMOTIC Dataset}.
\newblock {\em IEEE Trans. Pattern Anal. Mach. Intell.}, 42(11):2755--2766,
  2020.

\bibitem{kumar2020noisy}
V. Kumar, S. Rao, and L. Yu.
\newblock {Noisy Student Training Using Body Language Dataset Improves Facial
  Expression Recognition}.
\newblock In {\em ECCV Workshops}, pages 756--773. Springer, 2020.

\bibitem{li2019spatio}
B. Li, X. Li, Z. Zhang, and F. Wu.
\newblock {Spatio-Temporal Graph Routing for Skeleton-Based Action
  Recognition}.
\newblock In {\em AAAI-19}, volume~33, pages 8561--8568. AAAI Press, 2019.

\bibitem{lin2014microsoft}
T.~Y. Lin, M. Maire, S. Belongie, J. Hays, P. Perona, D. Ramanan, P.
  Doll{\'a}r, and C.~L. Zitnick.
\newblock {Microsoft COCO: Common Objects in Context}.
\newblock In {\em ECCV}, pages 740--755. Springer Cham, 2014.

\bibitem{luo2020arbee}
Y. Luo, J. Ye, R.~B. Adams, J. Li, M.~G. Newman, and J.~Z. Wang.
\newblock {ARBEE: Towards Automated Recognition of Bodily Expression of Emotion
  in the Wild}.
\newblock {\em {International Journal of Computer Vision}}, 128(1):1--25, 2020.

\bibitem{luvizon20182d}
D.~C. Luvizon, D. Picard, and H. Tabia.
\newblock {2D/3D Pose Estimation and Action Recognition Using Multitask Deep
  Learning}.
\newblock In {\em CVPR}, pages 5137--5146. IEEE, 2018.

\bibitem{mensink2014costa}
T. Mensink, E. Gavves, and C.~G.~M. Snoek.
\newblock {COSTA: Co-Occurrence Statistics for Zero-Shot Classification}.
\newblock In {\em CVPR}, pages 2441--2448. IEEE, 2014.

\bibitem{mollahosseini2016going}
A. Mollahosseini, D. Chan, and M.~H. Mahoor.
\newblock {Going deeper in facial expression recognition using deep neural
  networks}.
\newblock In {\em WACV}, pages 1--10. IEEE, 2016.

\bibitem{noroozi2018survey}
F. Noroozi, D. Kaminska, C. Corneanu, T. Sapinski, S. Escalera, and G.
  Anbarjafari.
\newblock {Survey on Emotional Body Gesture Recognition}.
\newblock {\em IEEE Trans. Affect. Comput.}, 12(2):505--523, 2021.

\bibitem{pons2020multitask}
G. Pons and D. Masip.
\newblock Multitask, multilabel, and multidomain learning with convolutional
  networks for emotion recognition.
\newblock {\em IEEE Trans. Cybern.}, 52(6):4764--4771, 2022.

\bibitem{poria2017review}
S. Poria, E. Cambria, R. Bajpai, and A. Hussain.
\newblock {A review of affective computing: From unimodal analysis to
  multimodal fusion}.
\newblock {\em Inf. Fusion}, 37:98--125, 2017.

\bibitem{redmon2016you}
J. Redmon, S. Divvala, R. Girshick, and A. Farhadi.
\newblock {You Only Look Once: Unified, Real-Time Object Detection}.
\newblock In {\em CVPR}, pages 779--788. IEEE, 2016.

\bibitem{schindler2008recognizing}
K. Schindler, L. Van~Gool, and B. De~Gelder.
\newblock {Recognizing emotions expressed by body pose: A biologically inspired
  neural model}.
\newblock {\em Neural Netw}, 21(9):1238--1246, 2008.

\bibitem{silverman1998density}
Bernard~W. Silverman.
\newblock {\em {Density Estimation for Statistics and Data Analysis}}.
\newblock Routledge, 1998.

\bibitem{simonyan2014two}
K. Simonyan and A. Zisserman.
\newblock {Two-Stream Convolutional Networks for Action Recognition in Videos}.
\newblock In {\em NIPS}, pages 568--576. MIT Press, 2014.

\bibitem{szegedy2015going}
C. Szegedy, W. Liu, Y. Jia, P. Sermanet, S. Reed, D. Anguelov, D. Erhan, V.
  Vanhoucke, and A. Rabinovich.
\newblock {Going deeper with convolutions}.
\newblock In {\em CVPR}, pages 1--9. IEEE, 2015.

\bibitem{tomei2021video}
M. Tomei, L. Baraldi, S. Calderara, S. Bronzin, and R. Cucchiara.
\newblock {Video action detection by learning graph-based spatio-temporal
  interactions}.
\newblock {\em Comput. Vis. Image Underst.}, 206:103187, 2021.

\bibitem{tran2018closer}
D. Tran, H. Wang, L. Torresani, J. Ray, Y. LeCun, and M. Paluri.
\newblock {A Closer Look at Spatiotemporal Convolutions for Action
  Recognition}.
\newblock In {\em CVPR}, pages 6450--6459. IEEE, 2018.

\bibitem{ulutan2020actor}
O. Ulutan, S. Rallapalli, M. Srivatsa, C. Torres, and B.~S. Manjunath.
\newblock {Actor Conditioned Attention Maps for Video Action Detection}.
\newblock In {\em WACV}, pages 516--525. IEEE, 2020.

\bibitem{von1966choreutics}
R. Von~Laban.
\newblock {\em Choreutics}.
\newblock Macdonald and Evans, 1966.

\bibitem{wang2016temporal}
L. Wang, Y. Xiong, Z. Wang, Yu Q., D. Lin, X. Tang, and L. Van~Gool.
\newblock {Temporal Segment Networks: Towards Good Practices for Deep Action
  Recognition}.
\newblock In {\em ECCV}, pages 20--36. Springer, 2016.

\bibitem{wang2018non}
X. Wang, R. Girshick, A. Gupta, and K. He.
\newblock Non-local neural networks.
\newblock In {\em CVPR}, pages 7794--7803, 2018.

\bibitem{xu2017microexpression}
F. Xu, J. Zhang, and J.~Z. Wang.
\newblock {Microexpression Identification and Categorization Using a Facial
  Dynamics Map}.
\newblock {\em IEEE Trans. Affect. Comput.}, 8(2):254--267, 2017.

\bibitem{yadav2021review}
S.~K. Yadav, K. Tiwari, H.~M. Pandey, and S.~A. Akbar.
\newblock {A review of multimodal human activity recognition with special
  emphasis on classification, applications, challenges and future directions}.
\newblock {\em Knowl. Based Syst.}, 223:106970, 2021.

\bibitem{yu2019unsupervised}
H.~X. Yu, W.~S. Zheng, A. Wu, X. Guo, S. Gong, and J.~H. Lai.
\newblock {Unsupervised Person Re-Identification by Soft Multilabel Learning}.
\newblock In {\em CVPR}, pages 2143--2152. IEEE, 2019.

\bibitem{zhou2018places}
B. Zhou, A. Lapedriza, A. Khosla, A. Oliva, and A. Torralba.
\newblock {Places: A 10 Million Image Database for Scene Recognition}.
\newblock {\em IEEE Trans. Pattern Anal. Mach. Intell.}, 40(6):1452--1464,
  2018.

\end{thebibliography}
}
\vspace*{-2\baselineskip}
\begin{IEEEbiography}[{\includegraphics[width=1in,height=1.25in,clip,keepaspectratio]{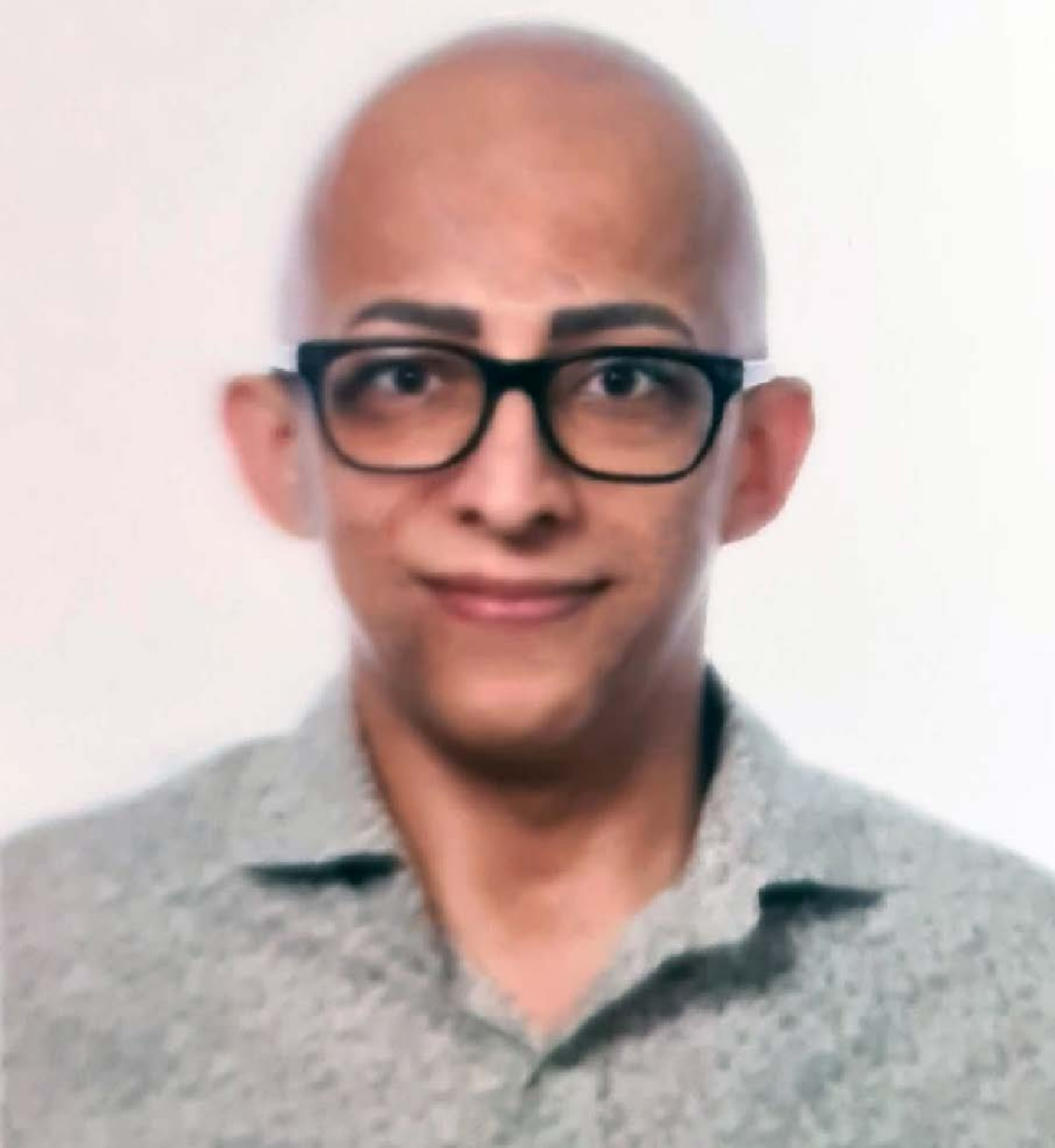}}]{Mohammad Mahdi Dehshibi} (Member, IEEE) received his PhD in Computer Science in 2017. He is currently a research scientist at Universidad Carlos III de Madrid, Spain. He is also an adjunct researcher at Universitat Oberta de Catalunya (Spain) and the Unconventional Computing Lab. at UWE (Bristol, UK). He has contributed to more than 70 papers published in peer-reviewed journals and conference proceedings. He also serves as an associate editor of the \emph{International Journal of Parallel, Emergent, and Distributed Systems}. His research interests include Deep Learning, Medical Image Processing, Human Behaviour Analysis, Unconventional Computing, and Affective Computing.
\end{IEEEbiography}
\vspace*{-2\baselineskip}
\begin{IEEEbiography}[{\includegraphics[width=1in,height=1.25in,clip,keepaspectratio]{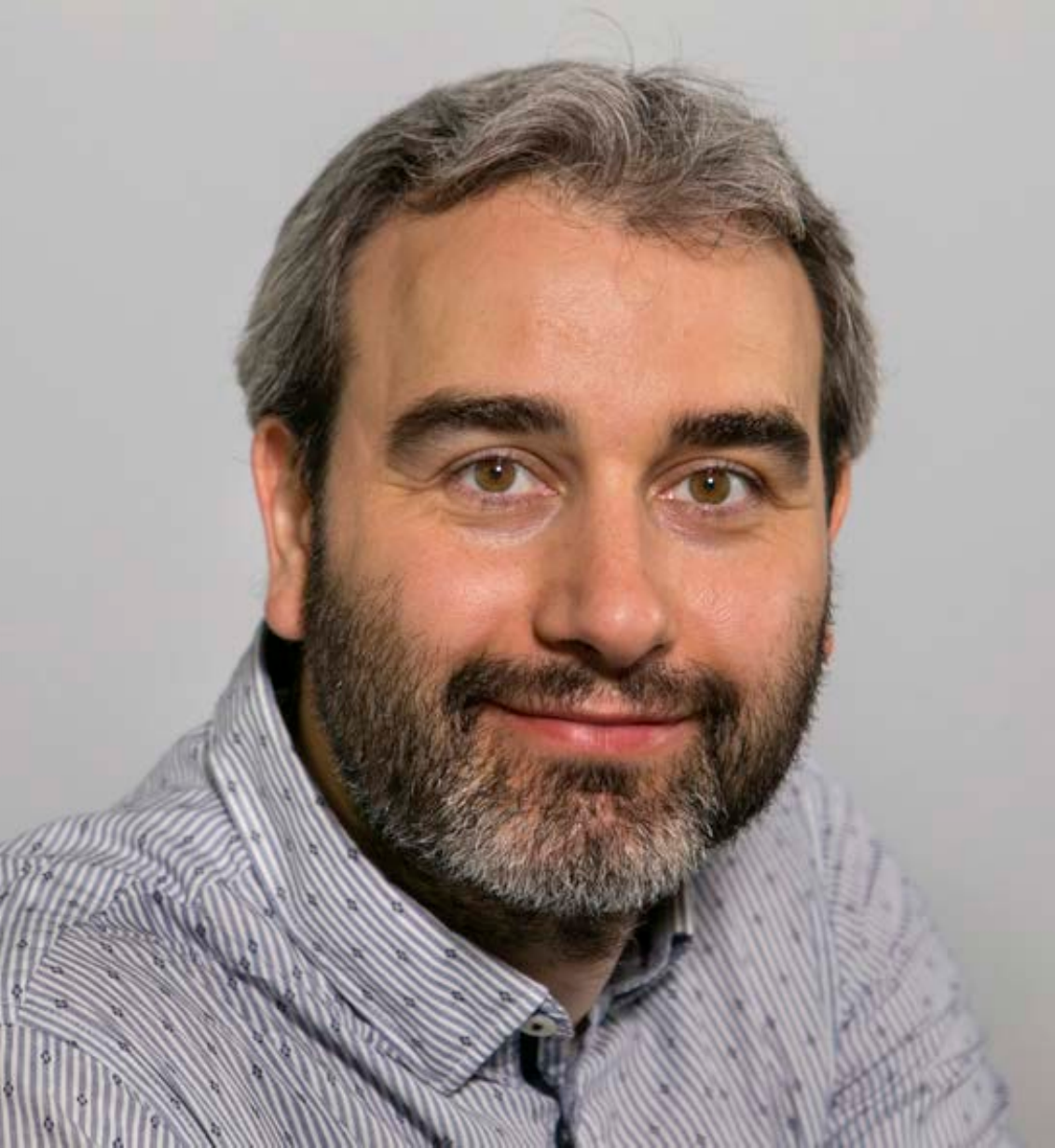}}]{David Masip} (Senior Member, IEEE) received his Ph.D. degree in Computer Vision in 2005 (Universitat Autonoma de Barcelona, Spain). He was awarded the best thesis in Computer Science. He is a Full Professor at the Computer Science, Multimedia, and Telecommunications Department at Universitat Oberta de Catalunya, Spain, and the Director of the Doctoral School since 2015. He has published more than 70 scientific papers in relevant journals and conferences. His research interests include Affective Computing, Oculomics, and Retina Image Analysis.
\end{IEEEbiography}
\vfill
\end{document}